\documentclass{article}

\usepackage{microtype}
\usepackage{graphicx}
\usepackage{subcaption}
\usepackage{booktabs} 
\usepackage{multirow}   
\usepackage{tabularx}   
\usepackage{amsmath}    

\usepackage{hyperref}

\usepackage[preprint]{icml2026}

\usepackage{amsmath}
\usepackage{amssymb}
\usepackage{mathtools}
\usepackage{amsthm}

\usepackage[capitalize,noabbrev]{cleveref}

\usepackage[most]{tcolorbox} 
\usepackage{afterpage} 
\usepackage{xurl} 

\theoremstyle{plain}

\theoremstyle{definition}

\theoremstyle{remark}

\usepackage[textsize=tiny]{todonotes}
\usepackage{makecell}

\icmltitlerunning{RooflineBench: A Benchmarking Framework for On-Device LLMs via Roofline Analysis}

\begin{document}
\twocolumn[
\icmltitle{RooflineBench: A Benchmarking Framework for \\ On-Device LLMs  via Roofline Analysis}



  \icmlsetsymbol{equal}{*}
  \icmlsetsymbol{corresponding}{$\dagger$}

  \begin{icmlauthorlist}
    \icmlauthor{Zhen Bi}{huzhou,equal}
    \icmlauthor{Xueshu Chen}{huzhou,equal}
    \icmlauthor{Luoyang Sun}{cas}
    \icmlauthor{Yuhang Yao}{cmu}
    \icmlauthor{Qing Shen}{huzhou}
    \icmlauthor{Jungang Lou}{huzhou,corresponding}
    \icmlauthor{Cheng Deng}{uk,corresponding}
  \end{icmlauthorlist}

  \icmlaffiliation{huzhou}{Huzhou University}
  \icmlaffiliation{cas}{Institution of Automation, Chinese Academy of Sciences}
  \icmlaffiliation{cmu}{  Carnegie Mellon University
  }
\icmlaffiliation{uk}{University of Edinburgh}
  \icmlcorrespondingauthor{Jungang Lou and Cheng Deng}{bizhen@zjhu.edu.cn  $|$ cdeng@ed.ac.uk}


  \icmlkeywords{On-devive LLMs, Benchmark}

  \vskip 0.3in
]



\printAffiliationsAndNotice{}  

\begin{abstract}
The transition toward localized intelligence through Small Language Models (SLMs) has intensified the need for rigorous performance characterization on resource-constrained edge hardware. However, objectively measuring the theoretical performance ceilings of diverse architectures across heterogeneous platforms remains a formidable challenge. In this work, we propose a systematic framework based on the Roofline model that unifies architectural primitives and hardware constraints through the lens of operational intensity ($OI$). By defining an inference-potential region, we introduce the \textit{Relative Inference Potential} as a novel metric to compare efficiency differences between Large Language Models (LLMs) on the same hardware substrate. Extensive empirical analysis across diverse compute tiers reveals that variations in performance and $OI$ are significantly influenced by sequence length. We further identify a critical regression in $OI$ as model depth increases. Additionally, our findings highlight an efficiency trap induced by hardware heterogeneity and demonstrate how structural refinements, such as Multi-head Latent Attention ($MLA$), can effectively unlock latent inference potential across various hardware substrates. These insights provide actionable directions for hardware-software co-design to align neural structures with physical constraints in on-device intelligence.
The released code is available in the Appendix~\ref{sec:Appendix C Experimental Details}.
\end{abstract}

\section{Introduction}





The rapid advancement of Large Language Models (LLMs) has redefined the landscape of artificial intelligence, yet their substantial scales continue to impose significant burdens on resource efficiency and deployment costs \cite{DBLP:journals/corr/abs-2203-15556}. This challenge has catalyzed a paradigm shift toward Small Language Models (SLMs) that prioritize capacity density and on-device accessibility \cite{DBLP:conf/icml/Liu0ILTFXCSKLC24}. Recent breakthroughs, including the Gemma, Phi-3, and MiniCPM families, demonstrate that compact architectures can rival significantly larger models while remaining deployable on consumer hardware \cite{DBLP:journals/corr/abs-2403-08295, DBLP:journals/corr/abs-2404-14219, DBLP:journals/corr/abs-2404-06395}. This trajectory further extends to the multimodal domain, where efficient models like MiniCPM-V and LLaVA-v1.6 surpass proprietary counterparts with minimal computational overhead \cite{DBLP:journals/corr/abs-2408-01800, yu2025minicpmv45cookingefficient, DBLP:conf/nips/LiWZULYNPG23}. Such developments underscore a critical transition toward localized intelligence that ensures data privacy and democratizes powerful AI capabilities across edge devices.

Recent literature has increasingly prioritized the quantitative assessment of inference efficiency through specialized metrics such as Model Bandwidth Utilization ($MBU$) \cite{databricks-mbu}, Model FLOPs Utilization ($MFU$) \cite{Chowdhery2022PaLMSL}, and evaluation frameworks like MoeCap \cite{jiang2025moecapbenchmarkingcostaccuracy}. Despite these advancements, establishing a comprehensive understanding of on-device intelligence remains a formidable task. Primarily, objectively characterizing the theoretical performance upper bound for specific large model architectures when deployed on heterogeneous hardware platforms is challenging because of the complex interplay between software kernels and hardware substrates \cite{dao2022flashattentionfastmemoryefficientexact}. Furthermore, conventional evaluation methods often lack the analytical depth required to pinpoint the fundamental physical constraints limiting inference efficacy in resource-constrained environments. These limitations necessitate a more robust analytical framework that can decompose hardware-software interactions to identify precise bottlenecks across diverse compute tiers.

\begin{figure*}[htbp]
    \centering
    \includegraphics[width=0.8\linewidth]{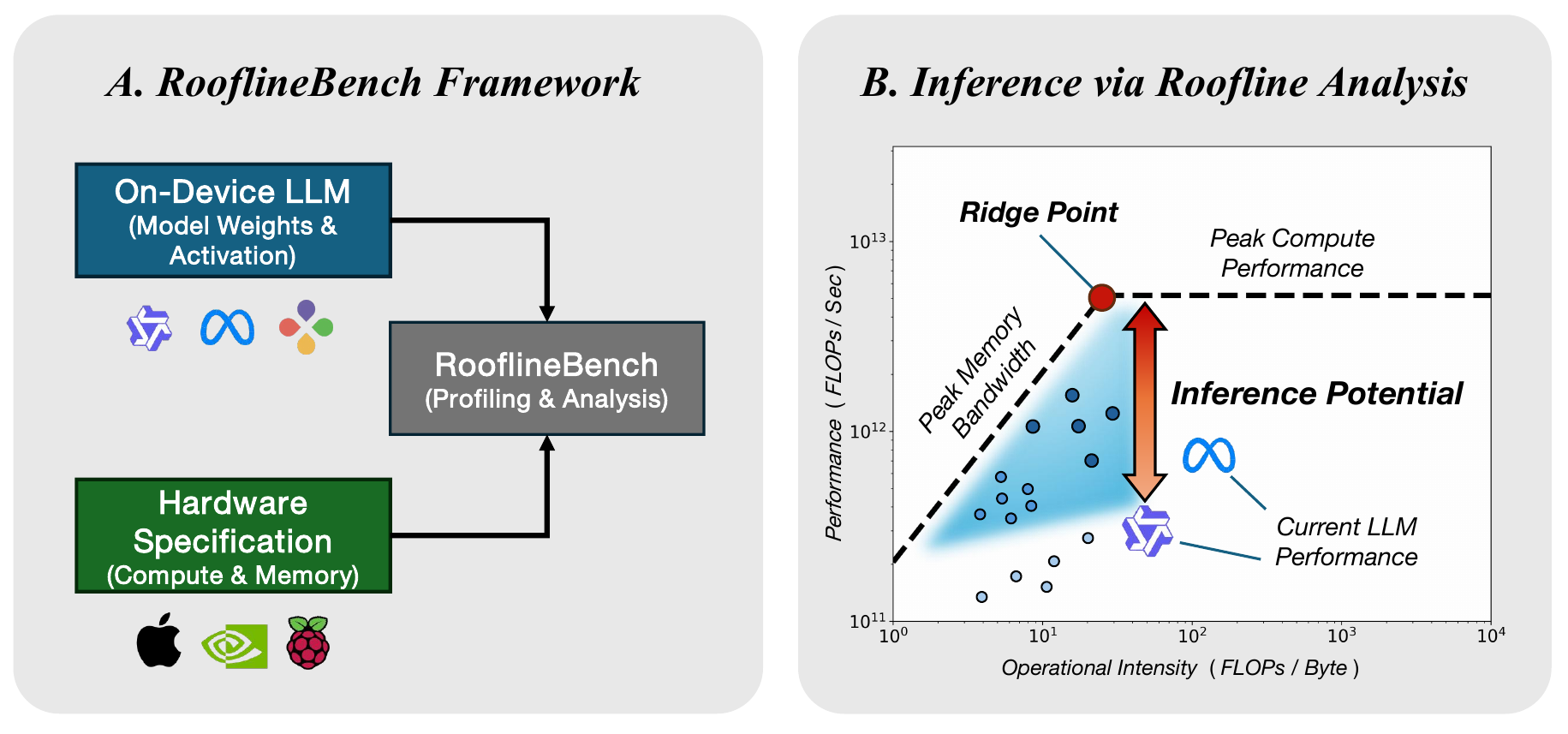}
    \caption{Overview of the RooflineBench framework and analytical methodology. 
    \textbf{Left:} The integrated profiling system incorporating on-device LLM configurations and hardware specifications for systematic analysis. 
    \textbf{Right:} Graphical representation of the Roofline model illustrating the \textit{Relative Inference Potential} between observed throughput and theoretical hardware ceilings across both memory-bound and compute-bound regimes.}
    \label{fig:main}
\end{figure*}

To address these challenges, we leverage the Roofline model as a rigorous analytical bridge to characterize the interplay between large model inference and underlying hardware capabilities \cite{DBLP:journals/cacm/WilliamsWP09}. Our approach entails the empirical measurement of peak $FLOPS$ and memory bandwidth to establish a realistic performance envelope across diverse platforms. 
We also argue that LLMs  exhibit an inference potential region on edge hardware devices (the blue region in in Figure \ref{fig:main}), and we propose a novel metric (\textit{Relative Inference Potential}) to compare the differences in inference potential between two LLMs on the same hardware device.
The primary contributions of this work are three fold:


\begin{itemize}

\item \textbf{Integrated Benchmarking Framework}: We propose a systematic Roofline based framework that unifies architectural primitives and hardware constraints via operational intensity ($OI$), defining an inference potential region to introduce the Relative Inference Potential ($\Phi$) for comparative efficiency analysis.

\item \textbf{Comprehensive Empirical Analysis}: Through extensive experiments across heterogeneous compute tiers, we reveal that inference efficiency is primarily governed by context length and attention architectures, while identifying a critical $OI$ regression at increased model depths.

\item \textbf{Hardware-Software Co-design Inspirations}: Our findings identify an efficiency trap induced by hardware heterogeneity and demonstrate how architectural refinements can maximize utilization across diverse hardware substrates to effectively bridge the gap between theoretical potential and real world execution.

\end{itemize}

\section{Background}

\paragraph{The Physics of On-Device Decoding}
Following the analytical framework of \textit{Pope et al.} \cite{DBLP:conf/mlsys/PopeDCDBHXAD23}, we deconstruct Transformer inference into two phases: prefill and decoding. While the prefill phase processes input tokens in parallel and is typically compute-bound, the decoding phase generates tokens autoregressively, where each step sequentially depends on previously generated tokens. This sequential nature inherently shifts the execution bottleneck toward memory bandwidth.

\paragraph{Mathematical Derivation of the Bandwidth Bottleneck}
For a decoder-only model with $n_{params}$ parameters, it has been shown that the forward pass requires approximately $2 \cdot n_{params}$ floating-point operations (FLOPs) per token . Let $P_{peak}$ be the peak performance of the hardware and $BW$ be the memory bandwidth. During decoding, the model weights and the attention key-value (KV) cache must be transferred from memory to compute cores for every single token. We define the \textit{compute time} ($T_{comp}$) and \textit{memory time} ($T_{mem}$) as:
\begin{align}
    T_{\text{comp}} &= \frac{2 \cdot n_{\text{params}}}{P_{\text{peak}}} \\
    T_{\text{mem}} &\approx \frac{n_{\text{params}} \cdot b_{\text{prec}}}{BW}
\end{align}
where $b_{prec}$ is the bytes per parameter. The system is memory-bandwidth bound when $T_{mem} > T_{comp}$. This constraint is captured by the \textit{Operational Intensity} (OI):
\begin{align}
    OI_{\text{dec}} \approx \frac{2 \cdot n_{\text{params}}}{n_{\text{params}} \cdot b_{\text{prec}}} = \frac{2}{b_{\text{prec}}}
\end{align}
For 16-bit precision ($b_{prec}=2$), $OI_{decode} \approx 1$ FLOP/Byte. Since modern edge accelerators often have $P_{peak}/BW$ ratios far exceeding 1, the computational cores remain essentially idle while waiting for tensors to be loaded, leading to low model FLOPs utilization (MFU).

\paragraph{Edge Heterogeneity and KV Cache Scaling}
On resource-constrained edge devices, the additional memory traffic from the KV cache further reduces the system's Operational Intensity (OI), shifting the execution point to the \textit{left} on the Roofline model \cite{DBLP:journals/cacm/WilliamsWP09}. Simultaneously, the limited memory bandwidth ($BW$) of mobile SoCs causes the hardware's ``knee point" ($P_{peak}/BW$) to shift to the \textit{right}. This dual effect forces the decoding phase deep into the severe memory-bandwidth bound region, where the computational cores remain underutilized while waiting for tensor loading. Consequently, architectural optimizations such as multiquery attention (MQA) are essential to reduce the KV cache footprint and alleviate this bandwidth pressure.



\section{Methodology: The Integrated Standard Roofline Framework}

This section presents our performance analysis framework. 
Unlike simulation-based approaches, our tool is runtime-integrated, designed to quantify LLM's inference potential by comparing real-time inference telemetry against empirically measured hardware limits.

\subsection{Standard Roofline Formulation}

We adopt the standard Roofline model to establish the theoretical performance upper bound. For a given hardware platform, attainable performance $P$ (GFLOPS) is bounded by either compute capacity or memory bandwidth:
$$
P = \min(P_{peak}, \ OI \times BW_{peak})
$$
Where
$P_{peak}$: Theoretical peak compute performance (GFLOPS).
$BW_{peak}$: Achievable peak memory bandwidth (GB/s).
$OI$ (Operational Intensity): The ratio of floating-point operations to bytes of memory traffic (FLOPs/Byte).
Given the autoregressive nature of LLM decoding (token-by-token generation), the process is inherently memory-bound. Consequently, our analysis focuses primarily on the \textbf{sloped region} of the Roofline graph, where performance is strictly limited by $OI \times BW_{peak}$.




\begin{figure*}[htbp]
    \centering
    \includegraphics[width=1.0\textwidth]{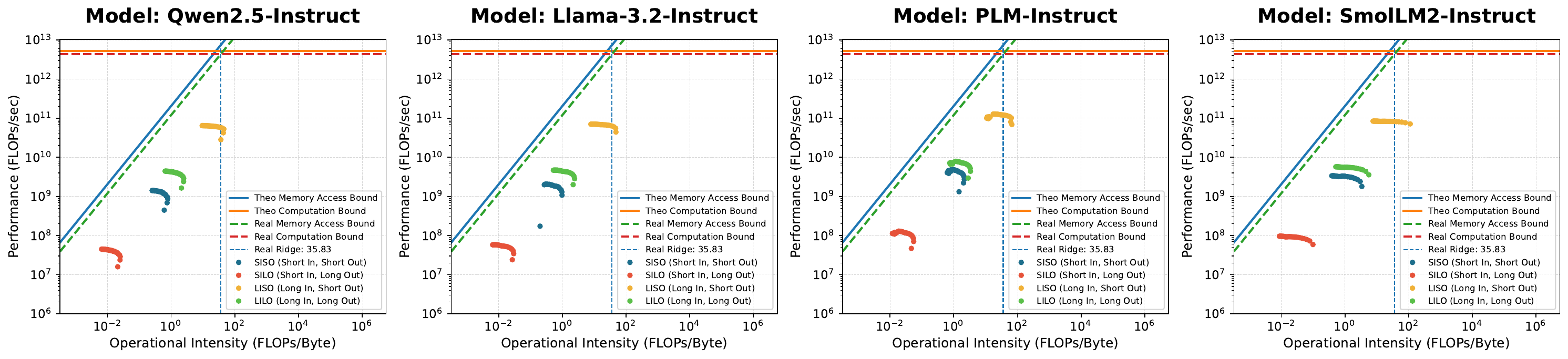}
    \caption{
    Comparison of roofline models across different input-output sequence length scenarios on Apple M1 Pro. Performance is evaluated in FP16 precision for Qwen2.5-Instruct (GQA), Llama-3.2-Instruct (GQA), PLM-Instruct (MLA), and SmolLM2-Instruct (MHA). The markers represent four inference scenarios: SISO, SILO, LISO, and LILO. Detailed analysis of the sensitivity to attention mechanisms is provided in subsequent sections. For comprehensive and detailed benchmarks across a wider range of hardware devices, please refer to Appendix \ref{sec:Appendix D Benchmark Results}.}
    \label{fig:Scenario_Comparison}
\end{figure*}

\subsection{Applying the Standard Roofline Model to LLMs}


To rigorous quantify the efficiency of LLM decoding, we instantiate the Roofline model by defining its three core coordinates: theoretical FLOPs, memory traffic, and empirical hardware limits.

\begin{itemize}
    
    \item \textbf{FLOPs ($W$): Analytical Estimation.}
    Directly measuring FLOPs via hardware counters can be imprecise due to instrumentation overhead. Instead, we adopt the analytical formulation from 
    \citet{DBLP:journals/corr/abs-2506-17286} to estimate the theoretical FLOPs for heterogeneous architectures (e.g., MHA, GQA).
    Given a model with hidden dimension $H$, sequence length $N$, and head configurations $\{n_q, n_k, n_v\}$, we precisely calculate the computational load for the Linear, Attention, and FFN layers per decoding step. This ensures a uniform standard for calculating Performance ($P$) and Operational Intensity ($OI$) across different backends.

    \item \textbf{Memory Traffic ($Q$): Data Movement Approximation.}
    In the context of memory-bound LLM decoding, memory traffic is dominated by loading model weights and the KV cache. We approximate the total data movement $Q$ per token generation as the summation of the model parameters size and the active KV cache entries read/written during that step. This approach captures the requisite data transfer volume regardless of the specific memory management implementation (e.g., Unified Memory on Apple Silicon vs. GDDR6X on NVIDIA GPUs).

    \item \textbf{Performance \& Operational Intensity Formulation.}
    Based on the derived $W$ and $Q$, combined with the measured end-to-end latency ($T$), we define the runtime metrics as:
    \begin{align}
        \text{Performance (GFLOPS)} &= \frac{W}{T} \\
        \text{Operational Intensity (FLOPs/Byte)} &= \frac{W}{Q}
    \end{align}
    
    \item \textbf{Hardware Limit Characterization.}
    We establish the theoretical ceiling by empirically profiling the peak memory bandwidth ($BW_{peak}$) and compute performance ($P_{peak}$).
    For Apple Silicon devices, we benchmark the Unified Memory bandwidth, as the GPU shares system memory. Conversely, for CUDA-enabled devices (e.g., RTX 4090), we isolate and measure the dedicated Video Memory (VRAM) bandwidth, as it constitutes the primary bottleneck for on-device inference.
\end{itemize}

\subsection{Characterizing Relative Inference Potential across Performance Regimes}


We define \textit{Relative Inference Potential} ($\Phi$) to quantify optimization headroom based on the spatial relationship between a performance point $P(OI_{p}, Perf_{p})$ and the hardware ridge $R(OI_{r}, \pi)$. The calculation logic varies across different regimes to reflect shifting physical constraints.

\textbf{Memory Bound Regime ($OI < OI_{r}$)}: $\Phi$ is the Euclidean distance to $R$, capturing the dual necessity of increasing operational intensity and throughput:
\begin{equation}\Phi(P) = \sqrt{(OI_{r} - OI_{p})^2 + (\pi - Perf_{p})^2}\end{equation}
\textbf{Compute Bound Regime ($OI \geq OI_{r}$)}: Since horizontal $OI$ gains yield negligible ceiling increases, $\Phi$ is defined as the vertical distance to the peak compute limit $\pi$:
\begin{equation}\Phi(P) = \pi - Perf_{p}\end{equation}
\textbf{Cross Regime Incomparability}: Data points on opposite sides of the ridge are fundamentally incomparable due to distinct physical bottlenecks (bandwidth vs computation), necessitating regime specific evaluation.

\section{Comprehensive Characterization Study}

This section provides a systematic characterization of LLM inference using the proposed \textit{Relative Inference Potential} ($\Phi$) within a Roofline analytical framework. Unlike traditional macro metrics such as throughput (TPS), $\Phi$ facilitates a regime aware evaluation of execution efficiency by quantifying the spatial discrepancy between realized performance and theoretical hardware limits. We evaluate task level sensitivity across four sequence patterns (Sec. \ref{sec:4.1 scenario analysis}), investigate parameter level scaling and bottleneck shifts through model depth profiling (Sec. \ref{sec:4.2 layers analysis}), and assess algorithmic impacts from precision and architectural refinements such as $MLA$ (Sec. \ref{sec:4.3 algorithmic analysis}).
\textbf{
By identifying critical transitions between memory bound and compute bound regimes, this multidimensional analysis leverages $\Phi$ to reveal the internal complexity and optimization headroom of localized intelligence. 
}
Experimental details are  in Appendix \ref{sec:Appendix C Experimental Details}.

\subsection{Sensitivity to Input-Output Sequence Lengths}\label{sec:4.1 scenario analysis}

To evaluate the impact of task-level workloads on on-device decoding efficiency, we categorize four representative sequence patterns based on typical edge-side use cases: (i) SISO (Short In, Short Out), such as local voice commands; (ii) SILO (Short In, Long Out), typical of creative writing or code completion; (iii) LISO (Long In, Short Out), reflecting RAG-based context extraction or document summarization; and (iv) LILO (Long In, Long Out), characteristic of document translation. These configurations fundamentally alter the ratio between attention-related computation and model weight loading during each autoregressive step.

\begin{figure*}[htbp]
    \centering
    \includegraphics[width=0.9\textwidth]{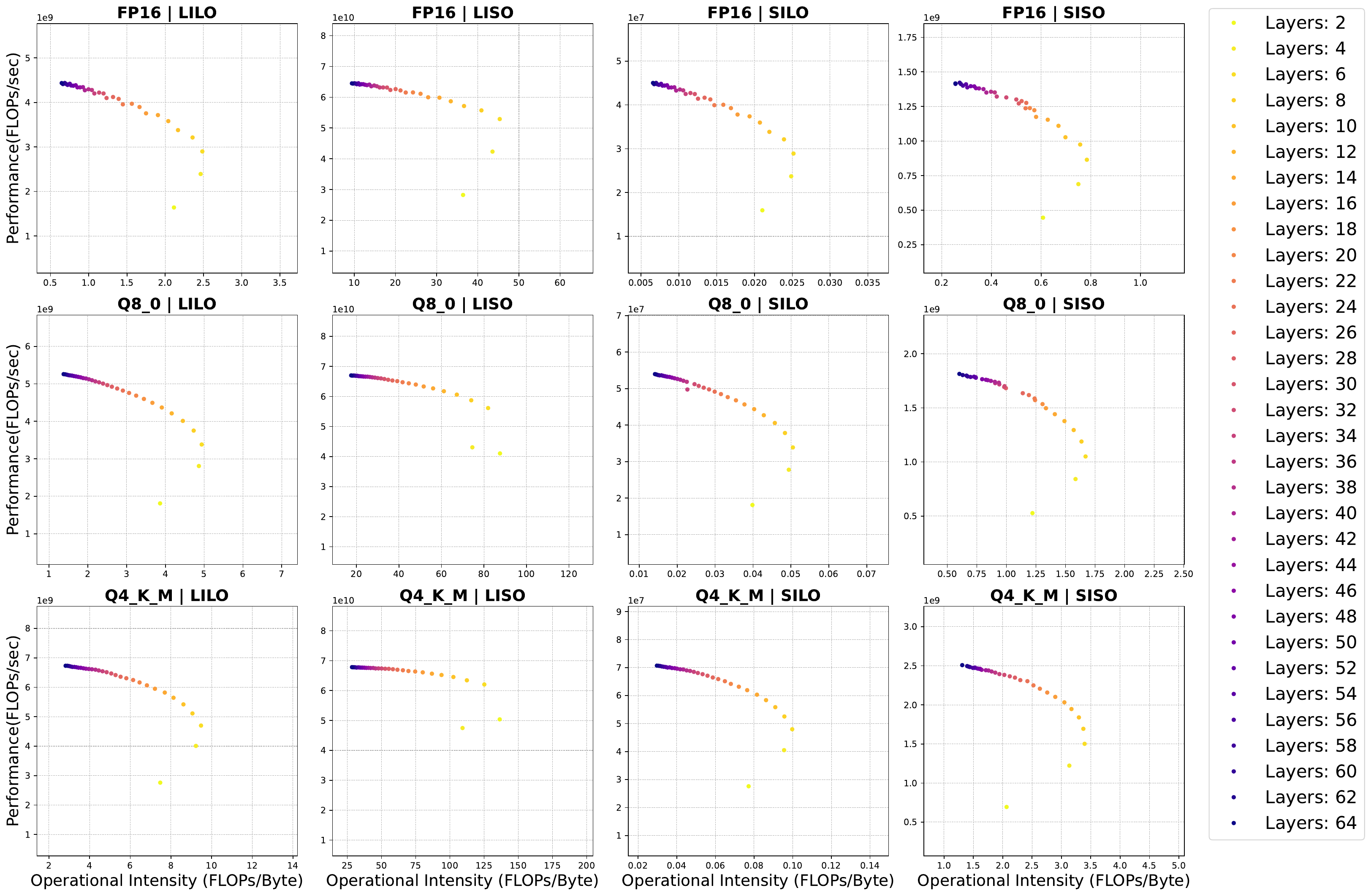}
    \caption{The figure shows the inference performance of Qwen2.5-Instruct FP16, Q8\_0, and Q4\_K\_M with 2 to 64 layers
    (Experimental platform: Apple M1 Pro).
    Comprehensive and detailed benchmarks across a wider range of hardware devices are in Appendix \ref{sec:Appendix D Benchmark Results}.}
    \label{fig:Qwen2.5-Instruct_all_scenarios}
\end{figure*}

In Figure \ref{fig:Scenario_Comparison}, decoding performance exhibits significant sensitivity to sequence lengths across all tested models. Notably, the LISO scenario (yellow dots) consistently achieves the highest execution efficiency, positioning itself as the closest observed case to the Roofline ridge point. This proximity stems from the larger input context, which increases the relative computational demand of the attention mechanism within each decoding step. By effectively amortizing the fixed memory overhead of loading model weights, LISO exhibits a high computational proportion, thereby elevating both the operational intensity ($OI$) and attainable performance ($GFLOPS$) toward the hardware's compute-bound limit. Conversely, the SILO scenario (red dots) resides deep within the memory-bound regime. In this scenario, the negligible computational requirement of processing a minimal context cannot offset the massive data movement of weights, leading to severe hardware underutilization.

\newtcolorbox{insight}{
    colback=blue!5!white,    
    colframe=white,          
    sharp corners,           
    boxrule=0pt,             
    left=5pt, right=5pt, top=5pt, bottom=5pt, 
    enhanced,                
}

\noindent
\begin{insight}
    \textbf{\textcolor{black}{Insight.1.}} \textit{Context length is the primary factor determining both the operational intensity and performance of on-device decoding: the LISO scenario approaches the compute-bound limit due to its inherently high computational proportion, while the SILO scenario remains severely memory-bound.}
\end{insight}

\subsection{Bottleneck Evolution under Parameter Expansion}\label{sec:4.2 layers analysis}

To investigate the impact of model scale on on-device inference dynamics, we evaluate a series of configurations by scaling the number of Transformer layers from 2 to 64. This range encompasses both ultra-lightweight edge models and standard mobile-class LLMs, providing a continuous spectrum to observe performance scaling laws across varying parameter capacities.

As illustrated in Figure \ref{fig:Qwen2.5-Instruct_all_scenarios}, scaling model depth reveals a distinct, non-linear evolution of hardware utilization on edge platforms. Initially, the attainable performance ($GFLOPS$) rises consistently and becomes increasingly dense as layers increase (transitioning from light yellow to dark purple dots). This behavior suggests that deeper models more effectively amortize fixed system-level overheads—such as kernel launch latencies, memory synchronization, and command orchestration—thereby enabling the hardware to operate closer to its steady-state compute limits.

More significantly, the operational intensity ($OI$) exhibits a non-monotonic "arch" trajectory with a clear inflection point at remarkably shallow depths. As depth increases from 2 to approximately 3--5 layers, the data points initially shift to the right, indicating an initial gain in arithmetic reuse as the system moves away from the overhead-dominated regime. However, beyond this threshold, the $OI$ begins to retreat to the left. In resource-constrained on-device environments, this regression arises because the cumulative memory bandwidth pressure of streaming weights for additional layers begins to outpace the marginal gains in computational reuse during the decoding phase. Consequently, the hardware "memory-wall" is encountered significantly earlier than theoretical predictions suggest, forcing the decoding process into an increasingly memory-bound regime as the model scales, which further constrains the release of relative inference-potential.

\begin{figure*}[htbp]
    \centering
    \includegraphics[width=0.84\textwidth]{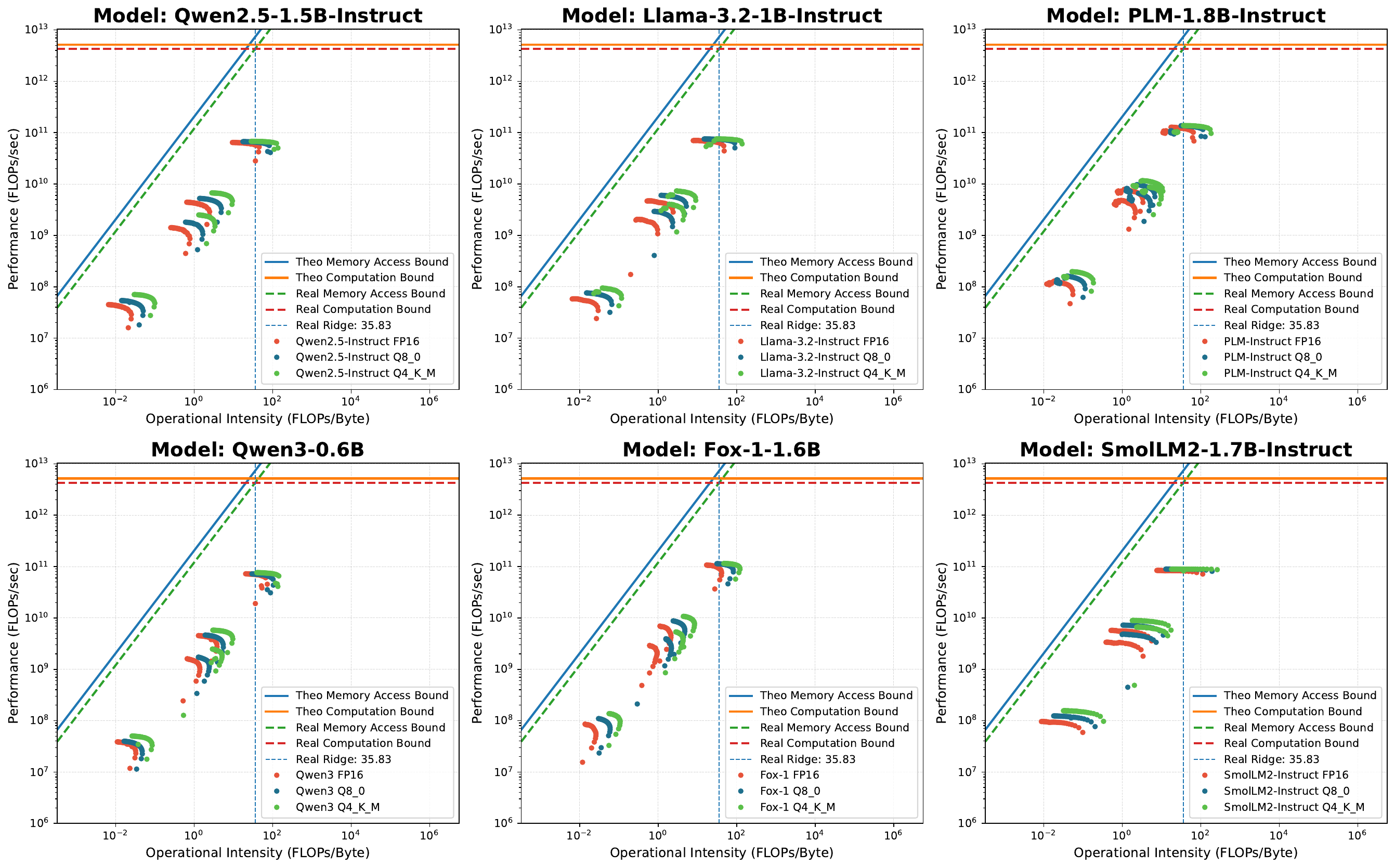}
    \caption{Performance comparison of various models across three different precision settings and all benchmarked hardware platforms. The results demonstrate a consistent performance gap and uniform trends, regardless of the specific device or precision level employed.}
    \label{fig:Different_Precision_Comparison}
    \vskip -4mm
\end{figure*}

\noindent
\begin{insight}
    \textbf{\textcolor{black}{Insight.2.}} \textit{On-device decoding reaches peak operational intensity at a remarkably shallow depth (3–5 layers), beyond which the memory bandwidth overhead of weight-streaming triggers a regression in operational intensity.}
\end{insight}

\subsection{Algorithmic Influences on Operational Intensity}\label{sec:4.3 algorithmic analysis}

Algorithmic optimizations fundamentally reshape the inference profile on the Roofline plane by modulating the balance between data movement and computation. This section evaluates the impact of numerical precision (Sec. \ref{sec:4.3.1 precision analysis}) and attention architectures (Sec. \ref{sec 4.3.2 attn type analysis}) on the operational intensity (OI) and bottleneck transitions of on-device LLMs.

\subsubsection{Numerical Precision: From FP16 to 4-bit Quantization}\label{sec:4.3.1 precision analysis}

To evaluate the impact of numerical precision on on-device efficiency, we compare FP16, Q8\_0, and Q\_K\_M formats. Quantization directly reduces the memory footprint of model weights, fundamentally reshaping the data movement characteristics during edge deployment. Detailed FLOPs calculation formulas for the KV cache and attention mechanisms across these architectures are in Appendix \ref{sec:Appendix C Experimental Details}.

In Figure \ref{fig:Different_Precision_Comparison}, lowering numerical precision triggers a diagonal shift of the data points toward the upper-right region of the Roofline plane. 
However, the magnitude of this improvement is highly sensitive to the task-level bottleneck.
In memory-bound scenarios such as SILO (bottom-left), transitioning from FP16 to Q4\_K\_M yields dramatic gains in both operational intensity (OI) and attainable performance ($GFLOPS$), as the reduced bandwidth pressure directly accelerates the execution. 
Conversely, in the LISO scenario (top-right), the clusters for the three precisions appear significantly tighter. 
Since LISO is already positioned near the Roofline ridge with high arithmetic intensity, the benefits of reduced data movement are less pronounced, indicating that the execution has largely transitioned from being memory-constrained to being limited by the hardware's peak compute capability.

\noindent
\begin{insight}
    \textbf{\textcolor{black}{Insight.3.}} \textit{Quantization provides maximal efficiency gains for memory-bound tasks (e.g., SILO), whereas in compute-heavy scenarios (e.g., LISO), the performance becomes increasingly saturated as it approaches the hardware's theoretical peak.}
\end{insight}

\subsubsection{Attention Mechanisms: Multi-Head vs. Sparse Variations}\label{sec 4.3.2 attn type analysis}

To investigate the impact of attention architectures on hardware efficiency, we compare three prevalent designs: 
Multi-Head Attention (MHA), Grouped-Query Attention (GQA) and Multi-head Latent Attention (MLA). To ensure fairness, all LLMs are scaled to a unified size of approximately 1.5B parameters by adjusting their respective layer counts.

In Figure \ref{fig:Different_Attention_Type_Model_Comparison}, the choice of attention mechanism significantly shifts the inference profile on the Roofline plane. MLA (blue dots) consistently achieves the highest operational intensity ($OI$) and attainable performance ($GFLOPS$) across all four sequence scenarios. By utilizing latent compression for the KV cache, MLA substantially reduces the volume of data movement required for each decoding step compared to the standard MHA, thereby shifting the execution significantly closer to the Roofline ridge point. Interestingly, at this specific 1.5B scale on edge hardware, GQA (red dots) exhibits the lowest efficiency among the three. It indicates that while GQA reduces the number of Key/Value heads, the latent-based compression in MLA provides a more effective balance between memory traffic reduction and computational throughput for on-device decoding.

\noindent
\begin{insight}
    \textbf{\textcolor{black}{Insight.4.}} \textit{Attention architecture is a decisive factor for decoding efficiency: MLA outperforms both MHA and GQA by effectively compressing KV cache traffic to maximize operational intensity on resource-constrained devices.}
\end{insight}

\begin{figure*}[htbp]
    \centering
    \includegraphics[width=0.8\textwidth]{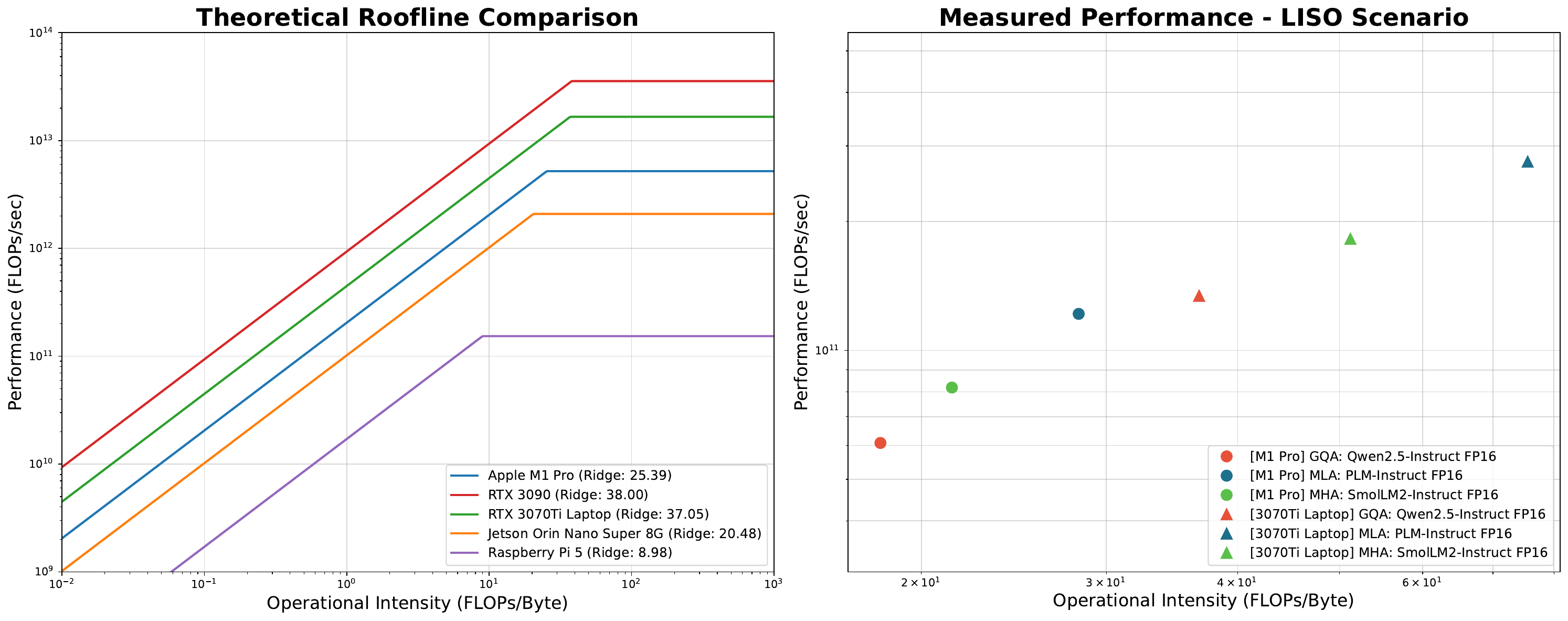
    }
    \caption{Analysis of hardware utilization efficiency and architectural scalability across heterogeneous platforms. The left panel presents the Theoretical Roofline Comparison for five representative devices including RTX 3090, RTX 3070Ti Laptop, Apple M1 Pro, Jetson Orin Nano, and Raspberry Pi 5, illustrating a significant disparity in theoretical ridge points ranging from 8.98 to 38.00 $FLOPs/Byte$. The right panel depicts the Measured Performance in the LISO scenario for MLA, MHA, and GQA configurations across the Apple M1 Pro and RTX 3070Ti Laptop, highlighting the consistent efficiency baseline and parallel shift in operational intensity and performance when scaling across hardware tiers.}
    \label{fig:Combined_Roofline_Comparison_LISO}
    \vskip -4mm
\end{figure*}

\section{A Fairness-View: Hardware Utilization Efficiency}

Hardware utilization efficiency is  constrained by the intrinsic disparities of heterogeneous platforms and the structural variations of model architectures.
This section evaluates the fairness of on-device inference by analyzing performance gaps across diverse hardware environments (Sec. \ref{sec:5.1 hardware}) and examining how architectural choices influence the equity of resource saturation at a unified model scale (Sec. \ref{sec:5.2 model arch}).

\subsection{Disparity across Heterogeneous Hardware Platforms}\label{sec:5.1 hardware}
Hardware utilization efficiency is strictly governed by the physical constraints of the underlying platform. As illustrated in Fig. \ref{fig:Combined_Roofline_Comparison_LISO}, the theoretical Roofline profiles of five representative devices, ranging from the high-performance RTX 3090 to the ultra-low-power Raspberry Pi 5, reveal a staggering disparity in hardware capabilities. The theoretical peak performance spans three orders of magnitude from $10^{11}$ to $10^{13}$ FLOPs/sec, while the memory bandwidth defines distinct operational boundaries for each platform.

The critical indicator of hardware unfairness is the theoretical ridge point, which dictates the minimum  $OI$ required to saturate computational resources. 
High-performance GPUs such as the RTX 3090 (Ridge: 38.00) and RTX 3070Ti Laptop (Ridge: 37.05) possess significantly higher ridge points, meaning they require a much higher $OI$ to transition from memory-bound to compute-bound regimes.
In contrast, edge-side chips like the Apple M1 Pro (Ridge: 25.39), Jetson Orin Nano (Ridge: 20.48) and Raspberry Pi 5 (Ridge: 8.98) have much lower ridge points. While these devices are easier to saturate, their absolute performance ceilings are considerably lower. This disparity implies that a specific model architecture, such as a 1.5B model in the LISO scenario, may achieve near-optimal saturation on a Jetson Orin or Raspberry Pi but remain severely underutilized on an RTX 3090 because of the structural differences in theoretical memory-to-compute ratios.

\noindent
\begin{insight}
    \textbf{\textcolor{black}{Insight.5.}} \textit{Hardware heterogeneity creates an "efficiency trap": the disparity in ridge points across devices ensures that a single model architecture cannot achieve uniform utilization equity, as the same operational intensity leads to vastly different bottleneck regimes on different platforms.}
\end{insight}

\subsection{Architectural Sensitivity of Utilization Equity}\label{sec:5.2 model arch}

\begin{figure}[htbp]
    \centering
    \includegraphics[width=0.4\textwidth]{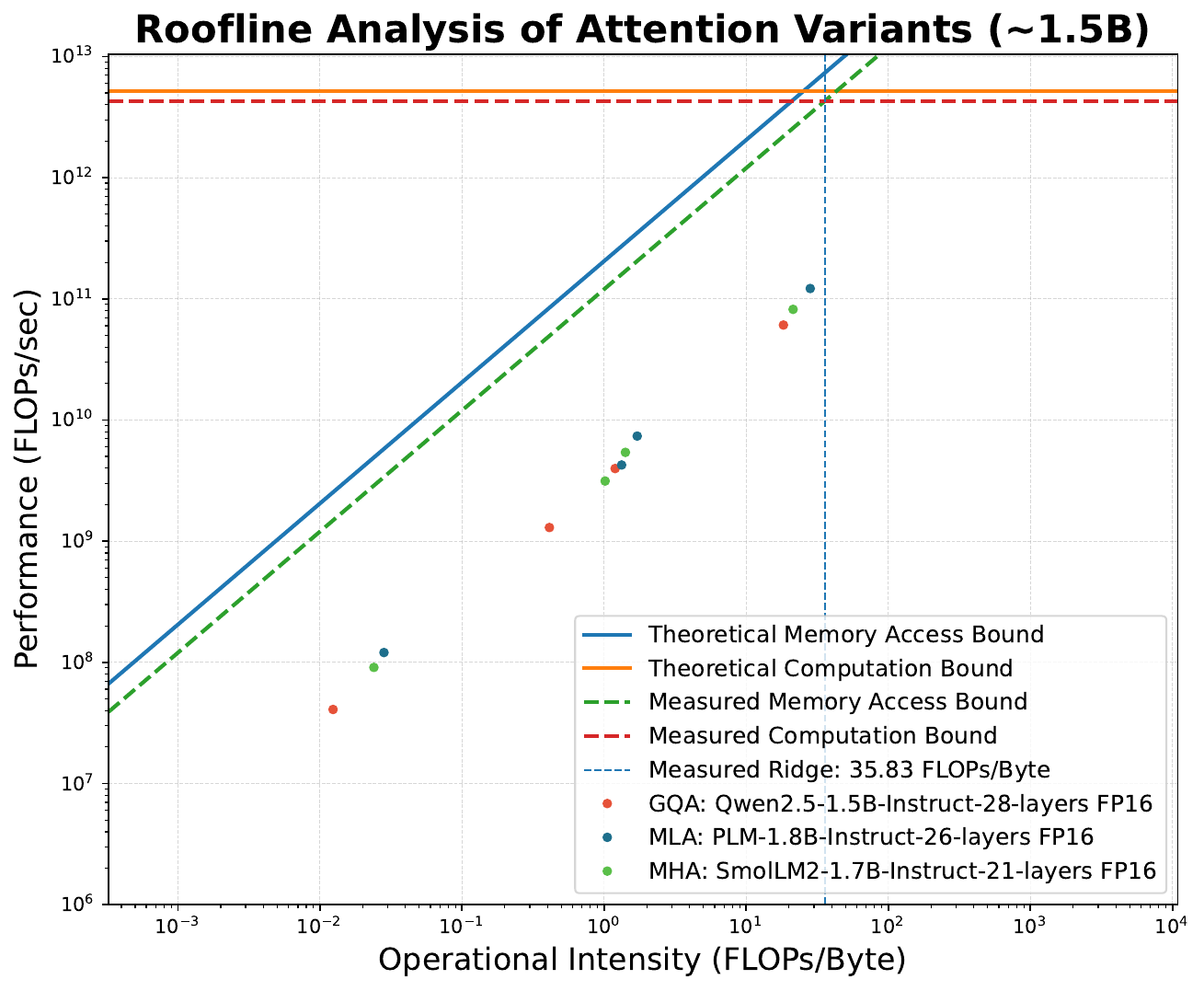}
    \caption{Impact of MHA, GQA, and MLA architectures on inference efficiency. All models are scaled to 1.5B parameters and tested at FP16 precision on an Apple M1 Pro. The distribution across the Roofline plane illustrates the efficiency gains of latent-based attention variations.}
    \label{fig:Different_Attention_Type_Model_Comparison}
    \vskip -4mm
\end{figure}

The cross-platform evaluation demonstrates that the efficiency dividends of architectural optimization are consistently preserved across varying hardware tiers. As illustrated in Fig. \ref{fig:Different_Attention_Type_Model_Comparison}, we analyze the performance of different architectural configurations including MLA, MHA, and GQA on the Apple M1 Pro and the RTX 3070Ti Laptop under the compute-intensive LISO scenario. The results indicate that as the execution environment scales from an edge SoC to a high-performance GPU, all architectural designs benefit from increased memory bandwidth and computational peaks, resulting in a parallel shift toward the upper-right region of the Roofline plane.

Despite the collective performance gains, the relative hierarchy of utilization equity remains sensitive to the specific architectural design. MLA (blue markers) consistently maintains a superior position in both operational intensity ($OI$) and attainable performance ($GFLOPS$) on both platforms. Rather than being a platform-specific enhancement, the latent-based compression in MLA functions as a robust efficiency baseline. This stability proves that optimized architectural structures can reliably capture available hardware resources across the device spectrum, ensuring that the performance advantages of high-end accelerators are not overshadowed by the structural inefficiencies of traditional attention mechanisms.

\noindent
\begin{insight}
    \textbf{\textcolor{black}{Insight.6.}} \textit{Architectural optimization demonstrates consistent cross-platform robustness because while hardware scaling elevates the performance of all configurations, optimized structures such as MLA maintain a higher baseline of operational intensity and attainable performance, ensuring superior utilization of heterogeneous computational resources.}
\end{insight}

\section{Discussion}

\paragraph{Synergy between Architectural Innovation and Hardware Constraints}

The architectural evolution toward Multi head Latent Attention (MLA), as seen in DeepSeek V2\cite{DBLP:journals/corr/abs-2405-04434} and PLM\cite{DBLP:journals/corr/abs-2503-12167}, along with the trainable sparse attention mechanism in MiniCPM\cite{DBLP:journals/corr/abs-2404-06395}, exemplifies the necessity of hardware aware innovation to transcend the memory wall in on device environments. 
MLA utilizes latent compression for the KV cache to reduce the volume of data movement required for each decoding step. Simultaneously, the sparse attention in MiniCPM reduces computational overhead by processing less than $5\%$ of tokens in long context scenarios. Both structural choices elevate the operational intensity ($OI$) and shift the inference profile toward the compute bound regime.
As shown in the cross platform comparison, MLA based configurations achieve a higher proficiency in capturing available computational peaks on high performance hardware compared to traditional mechanisms. 
Such synergy suggests that future on device architectures should prioritize optimizations such as latent compression or trainable sparsity to better align with the divergent theoretical ridge points of heterogeneous hardware.

\paragraph{Evolution of Capacity Density in Resource-Constrained Environments}

Capacity density, defined as the ratio of effective to actual parameter size, provides a unified framework for evaluating model quality according to the Densing Law\cite{DBLP:journals/corr/abs-2412-04315}. This metric is crucial for resource-constrained environments where traditional scaling through model size is increasingly unsustainable. Layer-wise analysis of models such as Qwen2.5\cite{DBLP:journals/corr/abs-2412-15115} reveals that operational intensity ($OI$) does not scale linearly with depth but reaches a peak at low layer counts. In the LISO scenario, increasing depth beyond this optimal threshold leads to an $OI$ fallback. Because edge platforms such as the Raspberry Pi 5 or Jetson Orin Nano have low theoretical ridge points, they are highly sensitive to $OI$ fluctuations. Consequently, improving the capacity density of shallow architectures is a more sustainable strategy for on-device intelligence than simply stacking layers, as it maximizes effective parameters within a restricted memory footprint.

\paragraph{Paradigm of Hardware-Targeted Specialization for Architectural Primitives}

The gap between theoretical peaks and realized throughput suggests that framework design must prioritize data movement over raw compute capacity. Although $TFLOPS$ have grown rapidly, inference remains fundamentally limited by memory access patterns across diverse devices. Refining the interaction between hierarchical memory structures and execution kernels is essential to mitigate the bandwidth constraints that govern large scale model deployment.
Additionally, hardware heterogeneity necessitates specializing AI compute units for specific high load primitives. General purpose logic frequently fails to exploit the structural advantages of modern architectures, leading to efficiency losses in on device scenarios. Hardware level optimization for critical primitives such as MLA and sparse attention is therefore crucial for maximizing throughput. Dedicated silicon support ensures that the theoretical operational intensity ($OI$) of optimized models can reach its full potential in real world execution.

\section{Conclusion}
We characterize LLM inference on edge devices using the Roofline model and introduce \textit{Relative Inference Potential} to compare efficiency across heterogeneous substrates. 
Findings show that sequence length dictates distinct performance and operational intensity ($OI$) dynamics.
We identify a non-monotonic $OI$ trajectory: performance initially rises via overhead amortization but regresses beyond 3--5 layers as parameter loading outpaces computational reuse. 
Our analysis reveals an efficiency trap from hardware ridge-point disparities, while structural refinements sustain higher $OI$ via optimized cache management.
These results advocate for hardware-software co-design to align model architectures with physical constraints and unlock localized intelligence.

\newpage


\bibliography{icml2026}

@article{DBLP:journals/corr/abs-2404-06395,
  author       = {Shengding Hu and
                  Yuge Tu and
                  Xu Han and
                  Chaoqun He and
                  Ganqu Cui and
                  Xiang Long and
                  Zhi Zheng and
                  Yewei Fang and
                  Yuxiang Huang and
                  Weilin Zhao and
                  Xinrong Zhang and
                  Zhen Leng Thai and
                  Kai Zhang and
                  Chongyi Wang and
                  Yuan Yao and
                  Chenyang Zhao and
                  Jie Zhou and
                  Jie Cai and
                  Zhongwu Zhai and
                  Ning Ding and
                  Chao Jia and
                  Guoyang Zeng and
                  Dahai Li and
                  Zhiyuan Liu and
                  Maosong Sun},
  title        = {MiniCPM: Unveiling the Potential of Small Language Models with Scalable
                  Training Strategies},
  journal      = {CoRR},
  volume       = {abs/2404.06395},
  year         = {2024},
  url          = {https://doi.org/10.48550/arXiv.2404.06395},
  doi          = {10.48550/ARXIV.2404.06395},
  eprinttype    = {arXiv},
  eprint       = {2404.06395},
  bibsource    = {dblp computer science bibliography, https://dblp.org}
}

@article{DBLP:journals/corr/abs-2203-15556,
  author       = {Jordan Hoffmann and
                  Sebastian Borgeaud and
                  Arthur Mensch and
                  Elena Buchatskaya and
                  Trevor Cai and
                  Eliza Rutherford and
                  Diego de Las Casas and
                  Lisa Anne Hendricks and
                  Johannes Welbl and
                  Aidan Clark and
                  Tom Hennigan and
                  Eric Noland and
                  Katie Millican and
                  George van den Driessche and
                  Bogdan Damoc and
                  Aurelia Guy and
                  Simon Osindero and
                  Karen Simonyan and
                  Erich Elsen and
                  Jack W. Rae and
                  Oriol Vinyals and
                  Laurent Sifre},
  title        = {Training Compute-Optimal Large Language Models},
  journal      = {CoRR},
  volume       = {abs/2203.15556},
  year         = {2022},
  url          = {https://doi.org/10.48550/arXiv.2203.15556},
  doi          = {10.48550/ARXIV.2203.15556},
  eprinttype    = {arXiv},
  eprint       = {2203.15556},
  bibsource    = {dblp computer science bibliography, https://dblp.org}
}

@inproceedings{DBLP:conf/icml/Liu0ILTFXCSKLC24,
  author       = {Zechun Liu and
                  Changsheng Zhao and
                  Forrest N. Iandola and
                  Chen Lai and
                  Yuandong Tian and
                  Igor Fedorov and
                  Yunyang Xiong and
                  Ernie Chang and
                  Yangyang Shi and
                  Raghuraman Krishnamoorthi and
                  Liangzhen Lai and
                  Vikas Chandra},
  title        = {MobileLLM: Optimizing Sub-billion Parameter Language Models for On-Device
                  Use Cases},
  booktitle    = {Forty-first International Conference on Machine Learning, {ICML} 2024,
                  Vienna, Austria, July 21-27, 2024},
  year         = {2024},
  url          = {https://openreview.net/forum?id=EIGbXbxcUQ},
  bibsource    = {dblp computer science bibliography, https://dblp.org}
}

@misc{yu2025minicpmv45cookingefficient,
      title={MiniCPM-V 4.5: Cooking Efficient MLLMs via Architecture, Data, and Training Recipe}, 
      author={Tianyu Yu and Zefan Wang and Chongyi Wang and Fuwei Huang and Wenshuo Ma and Zhihui He and Tianchi Cai and Weize Chen and Yuxiang Huang and Yuanqian Zhao and Bokai Xu and Junbo Cui and Yingjing Xu and Liqing Ruan and Luoyuan Zhang and Hanyu Liu and Jingkun Tang and Hongyuan Liu and Qining Guo and Wenhao Hu and Bingxiang He and Jie Zhou and Jie Cai and Ji Qi and Zonghao Guo and Chi Chen and Guoyang Zeng and Yuxuan Li and Ganqu Cui and Ning Ding and Xu Han and Yuan Yao and Zhiyuan Liu and Maosong Sun},
      year={2025},
      eprint={2509.18154},
      archivePrefix={arXiv},
      primaryClass={cs.LG},
      url={https://arxiv.org/abs/2509.18154}, 
}

@article{DBLP:journals/corr/abs-2403-08295,
  author       = {Gemma Team},
  title        = {Gemma: Open Models Based on Gemini Research and Technology},
  journal      = {CoRR},
  volume       = {abs/2403.08295},
  year         = {2024},
  url          = {https://doi.org/10.48550/arXiv.2403.08295},
  doi          = {10.48550/ARXIV.2403.08295},
  eprinttype    = {arXiv},
  eprint       = {2403.08295},
  bibsource    = {dblp computer science bibliography, https://dblp.org}
}

@article{DBLP:journals/corr/abs-2404-14219,
  author       = {Marah I Abdin and
                  Sam Ade Jacobs and
                  Ammar Ahmad Awan and
                  Jyoti Aneja and
                  Ahmed Awadallah and
                  Hany Awadalla and
                  Nguyen Bach and
                  Amit Bahree and
                  Arash Bakhtiari and
                  Harkirat S. Behl and
                  Alon Benhaim and
                  Misha Bilenko and
                  Johan Bjorck and
                  S{\'{e}}bastien Bubeck and
                  Martin Cai and
                  Caio C{\'{e}}sar Teodoro Mendes and
                  Weizhu Chen and
                  Vishrav Chaudhary and
                  Parul Chopra and
                  Allie Del Giorno and
                  Gustavo de Rosa and
                  Matthew Dixon and
                  Ronen Eldan and
                  Dan Iter and
                  Amit Garg and
                  Abhishek Goswami and
                  Suriya Gunasekar and
                  Emman Haider and
                  Junheng Hao and
                  Russell J. Hewett and
                  Jamie Huynh and
                  Mojan Javaheripi and
                  Xin Jin and
                  Piero Kauffmann and
                  Nikos Karampatziakis and
                  Dongwoo Kim and
                  Mahoud Khademi and
                  Lev Kurilenko and
                  James R. Lee and
                  Yin Tat Lee and
                  Yuanzhi Li and
                  Chen Liang and
                  Weishung Liu and
                  Eric Lin and
                  Zeqi Lin and
                  Piyush Madan and
                  Arindam Mitra and
                  Hardik Modi and
                  Anh Nguyen and
                  Brandon Norick and
                  Barun Patra and
                  Daniel Perez{-}Becker and
                  Thomas Portet and
                  Reid Pryzant and
                  Heyang Qin and
                  Marko Radmilac and
                  Corby Rosset and
                  Sambudha Roy and
                  Olatunji Ruwase and
                  Olli Saarikivi and
                  Amin Saied and
                  Adil Salim and
                  Michael Santacroce and
                  Shital Shah and
                  Ning Shang and
                  Hiteshi Sharma and
                  Xia Song and
                  Masahiro Tanaka and
                  Xin Wang and
                  Rachel Ward and
                  Guanhua Wang and
                  Philipp A. Witte and
                  Michael Wyatt and
                  Can Xu and
                  Jiahang Xu and
                  Sonali Yadav and
                  Fan Yang and
                  Ziyi Yang and
                  Donghan Yu and
                  Chengruidong Zhang and
                  Cyril Zhang and
                  Jianwen Zhang and
                  Li Lyna Zhang and
                  Yi Zhang and
                  Yue Zhang and
                  Yunan Zhang and
                  Xiren Zhou},
  title        = {Phi-3 Technical Report: {A} Highly Capable Language Model Locally
                  on Your Phone},
  journal      = {CoRR},
  volume       = {abs/2404.14219},
  year         = {2024},
  url          = {https://doi.org/10.48550/arXiv.2404.14219},
  doi          = {10.48550/ARXIV.2404.14219},
  eprinttype    = {arXiv},
  eprint       = {2404.14219},
  bibsource    = {dblp computer science bibliography, https://dblp.org}
}

@article{DBLP:journals/corr/abs-2408-01800,
  author       = {Yuan Yao and
                  Tianyu Yu and
                  Ao Zhang and
                  Chongyi Wang and
                  Junbo Cui and
                  Hongji Zhu and
                  Tianchi Cai and
                  Haoyu Li and
                  Weilin Zhao and
                  Zhihui He and
                  Qianyu Chen and
                  Huarong Zhou and
                  Zhensheng Zou and
                  Haoye Zhang and
                  Shengding Hu and
                  Zhi Zheng and
                  Jie Zhou and
                  Jie Cai and
                  Xu Han and
                  Guoyang Zeng and
                  Dahai Li and
                  Zhiyuan Liu and
                  Maosong Sun},
  title        = {MiniCPM-V: {A} {GPT-4V} Level {MLLM} on Your Phone},
  journal      = {CoRR},
  volume       = {abs/2408.01800},
  year         = {2024},
  url          = {https://doi.org/10.48550/arXiv.2408.01800},
  doi          = {10.48550/ARXIV.2408.01800},
  eprinttype    = {arXiv},
  eprint       = {2408.01800},
  bibsource    = {dblp computer science bibliography, https://dblp.org}
}

@inproceedings{DBLP:conf/nips/LiWZULYNPG23,
  author       = {Chunyuan Li and
                  Cliff Wong and
                  Sheng Zhang and
                  Naoto Usuyama and
                  Haotian Liu and
                  Jianwei Yang and
                  Tristan Naumann and
                  Hoifung Poon and
                  Jianfeng Gao},
  editor       = {Alice Oh and
                  Tristan Naumann and
                  Amir Globerson and
                  Kate Saenko and
                  Moritz Hardt and
                  Sergey Levine},
  title        = {LLaVA-Med: Training a Large Language-and-Vision Assistant for Biomedicine
                  in One Day},
  booktitle    = {Advances in Neural Information Processing Systems 36: Annual Conference
                  on Neural Information Processing Systems 2023, NeurIPS 2023, New Orleans,
                  LA, USA, December 10 - 16, 2023},
  year         = {2023},
  url          = {http://papers.nips.cc/paper\_files/paper/2023/hash/5abcdf8ecdcacba028c6662789194572-Abstract-Datasets\_and\_Benchmarks.html},
  bibsource    = {dblp computer science bibliography, https://dblp.org}
}

@misc{databricks-mbu,
  author = {Megha Agarwal
            and Asfandyar Qureshi
            and Nikhil Sardana
            and Linden Li
            and Julian Quevedo
            and Daya Khudia},
  year = {2023},
  publisher = {{Databricks}},
  howpublished = "\url{https://www.databricks.com/blog/llm-inference-performance-engineering-best-practices}"
}

@article{Chowdhery2022PaLMSL,
  author       = {Aakanksha Chowdhery and
                  Sharan Narang and
                  Jacob Devlin and
                  Maarten Bosma and
                  Gaurav Mishra and
                  Adam Roberts and
                  Paul Barham and
                  Hyung Won Chung and
                  Charles Sutton and
                  Sebastian Gehrmann and
                  Parker Schuh and
                  Kensen Shi and
                  Sasha Tsvyashchenko and
                  Joshua Maynez and
                  Abhishek Rao and
                  Parker Barnes and
                  Yi Tay and
                  Noam Shazeer and
                  Vinodkumar Prabhakaran and
                  Emily Reif and
                  Nan Du and
                  Ben Hutchinson and
                  Reiner Pope and
                  James Bradbury and
                  Jacob Austin and
                  Michael Isard and
                  Guy Gur{-}Ari and
                  Pengcheng Yin and
                  Toju Duke and
                  Anselm Levskaya and
                  Sanjay Ghemawat and
                  Sunipa Dev and
                  Henryk Michalewski and
                  Xavier Garcia and
                  Vedant Misra and
                  Kevin Robinson and
                  Liam Fedus and
                  Denny Zhou and
                  Daphne Ippolito and
                  David Luan and
                  Hyeontaek Lim and
                  Barret Zoph and
                  Alexander Spiridonov and
                  Ryan Sepassi and
                  David Dohan and
                  Shivani Agrawal and
                  Mark Omernick and
                  Andrew M. Dai and
                  Thanumalayan Sankaranarayana Pillai and
                  Marie Pellat and
                  Aitor Lewkowycz and
                  Erica Moreira and
                  Rewon Child and
                  Oleksandr Polozov and
                  Katherine Lee and
                  Zongwei Zhou and
                  Xuezhi Wang and
                  Brennan Saeta and
                  Mark Diaz and
                  Orhan Firat and
                  Michele Catasta and
                  Jason Wei and
                  Kathy Meier{-}Hellstern and
                  Douglas Eck and
                  Jeff Dean and
                  Slav Petrov and
                  Noah Fiedel},
  title        = {{PaLM}: Scaling Language Modeling with Pathways},
  journal      = {J. Mach. Learn. Res.},
  volume       = {24},
  pages        = {240:1--240:113},
  year         = {2023}
}

@misc{dao2022flashattentionfastmemoryefficientexact,
      title={FlashAttention: Fast and Memory-Efficient Exact Attention with IO-Awareness}, 
      author={Tri Dao and Daniel Y. Fu and Stefano Ermon and Atri Rudra and Christopher Ré},
      year={2022},
      eprint={2205.14135},
      archivePrefix={arXiv},
      primaryClass={cs.LG},
      url={https://arxiv.org/abs/2205.14135}, 
}

@inproceedings{DBLP:conf/mlsys/PopeDCDBHXAD23,
  author       = {Reiner Pope and
                  Sholto Douglas and
                  Aakanksha Chowdhery and
                  Jacob Devlin and
                  James Bradbury and
                  Jonathan Heek and
                  Kefan Xiao and
                  Shivani Agrawal and
                  Jeff Dean},
  editor       = {Dawn Song and
                  Michael Carbin and
                  Tianqi Chen},
  title        = {Efficiently Scaling Transformer Inference},
  booktitle    = {Proceedings of the Sixth Conference on Machine Learning and Systems,
                  MLSys 2023, Miami, FL, USA, June 4-8, 2023},
  publisher    = {mlsys.org},
  year         = {2023},
  url          = {https://proceedings.mlsys.org/paper\_files/paper/2023/hash/c4be71ab8d24cdfb45e3d06dbfca2780-Abstract-mlsys2023.html},
  bibsource    = {dblp computer science bibliography, https://dblp.org}
}

@article{DBLP:journals/corr/abs-2506-17286,
  author       = {Luoyang Sun and
                  Jiwen Jiang and
                  Cheng Deng and
                  Xinjian Wu and
                  Haifeng Zhang and
                  Lei Chen and
                  Lionel M. Ni and
                  Jun Wang},
  title        = {{GTA:} Grouped-head latenT Attention},
  journal      = {CoRR},
  volume       = {abs/2506.17286},
  year         = {2025},
  url          = {https://doi.org/10.48550/arXiv.2506.17286},
  doi          = {10.48550/ARXIV.2506.17286},
  eprinttype    = {arXiv},
  eprint       = {2506.17286},
  bibsource    = {dblp computer science bibliography, https://dblp.org}
}

@article{DBLP:journals/corr/abs-2412-04315,
  author       = {Chaojun Xiao and
                  Jie Cai and
                  Weilin Zhao and
                  Guoyang Zeng and
                  Biyuan Lin and
                  Jie Zhou and
                  Zhi Zheng and
                  Xu Han and
                  Zhiyuan Liu and
                  Maosong Sun},
  title        = {Densing Law of LLMs},
  journal      = {CoRR},
  volume       = {abs/2412.04315},
  year         = {2024},
  url          = {https://doi.org/10.48550/arXiv.2412.04315},
  doi          = {10.48550/ARXIV.2412.04315},
  eprinttype    = {arXiv},
  eprint       = {2412.04315},
  bibsource    = {dblp computer science bibliography, https://dblp.org}
}

@article{DBLP:journals/corr/abs-2405-04434,
  author       = {DeepSeek{-}AI},
  title        = {DeepSeek-V2: {A} Strong, Economical, and Efficient Mixture-of-Experts
                  Language Model},
  journal      = {CoRR},
  volume       = {abs/2405.04434},
  year         = {2024},
  url          = {https://doi.org/10.48550/arXiv.2405.04434},
  doi          = {10.48550/ARXIV.2405.04434},
  eprinttype    = {arXiv},
  eprint       = {2405.04434},
  bibsource    = {dblp computer science bibliography, https://dblp.org}
}

@article{DBLP:journals/corr/abs-2503-12167,
  author       = {Cheng Deng and
                  Luoyang Sun and
                  Jiwen Jiang and
                  Yongcheng Zeng and
                  Xinjian Wu and
                  Wenxin Zhao and
                  Qingfa Xiao and
                  Jiachuan Wang and
                  Haoyang Li and
                  Lei Chen and
                  Lionel M. Ni and
                  Haifeng Zhang and
                  Jun Wang},
  title        = {{PLM:} Efficient Peripheral Language Models Hardware-Co-Designed for
                  Ubiquitous Computing},
  journal      = {CoRR},
  volume       = {abs/2503.12167},
  year         = {2025},
  url          = {https://doi.org/10.48550/arXiv.2503.12167},
  doi          = {10.48550/ARXIV.2503.12167},
  eprinttype    = {arXiv},
  eprint       = {2503.12167},
  bibsource    = {dblp computer science bibliography, https://dblp.org}
}

@article{DBLP:journals/corr/abs-2412-15115,
  author       = {An Yang and
                  Baosong Yang and
                  Beichen Zhang and
                  Binyuan Hui and
                  Bo Zheng and
                  Bowen Yu and
                  Chengyuan Li and
                  Dayiheng Liu and
                  Fei Huang and
                  Haoran Wei and
                  Huan Lin and
                  Jian Yang and
                  Jianhong Tu and
                  Jianwei Zhang and
                  Jianxin Yang and
                  Jiaxi Yang and
                  Jingren Zhou and
                  Junyang Lin and
                  Kai Dang and
                  Keming Lu and
                  Keqin Bao and
                  Kexin Yang and
                  Le Yu and
                  Mei Li and
                  Mingfeng Xue and
                  Pei Zhang and
                  Qin Zhu and
                  Rui Men and
                  Runji Lin and
                  Tianhao Li and
                  Tingyu Xia and
                  Xingzhang Ren and
                  Xuancheng Ren and
                  Yang Fan and
                  Yang Su and
                  Yichang Zhang and
                  Yu Wan and
                  Yuqiong Liu and
                  Zeyu Cui and
                  Zhenru Zhang and
                  Zihan Qiu},
  title        = {Qwen2.5 Technical Report},
  journal      = {CoRR},
  volume       = {abs/2412.15115},
  year         = {2024},
  url          = {https://doi.org/10.48550/arXiv.2412.15115},
  doi          = {10.48550/ARXIV.2412.15115},
  eprinttype    = {arXiv},
  eprint       = {2412.15115},
  bibsource    = {dblp computer science bibliography, https://dblp.org}
}

@article{DBLP:journals/corr/abs-2001-08361,
  author       = {Jared Kaplan and
                  Sam McCandlish and
                  Tom Henighan and
                  Tom B. Brown and
                  Benjamin Chess and
                  Rewon Child and
                  Scott Gray and
                  Alec Radford and
                  Jeffrey Wu and
                  Dario Amodei},
  title        = {Scaling Laws for Neural Language Models},
  journal      = {CoRR},
  volume       = {abs/2001.08361},
  year         = {2020},
  url          = {https://arxiv.org/abs/2001.08361},
  eprinttype    = {arXiv},
  eprint       = {2001.08361},
  bibsource    = {dblp computer science bibliography, https://dblp.org}
}

@article{DBLP:journals/corr/abs-2307-09288,
  author       = {Hugo Touvron and
                  Louis Martin and
                  Kevin Stone and
                  Peter Albert and
                  Amjad Almahairi and
                  Yasmine Babaei and
                  Nikolay Bashlykov and
                  Soumya Batra and
                  Prajjwal Bhargava and
                  Shruti Bhosale and
                  Dan Bikel and
                  Lukas Blecher and
                  Cristian Canton{-}Ferrer and
                  Moya Chen and
                  Guillem Cucurull and
                  David Esiobu and
                  Jude Fernandes and
                  Jeremy Fu and
                  Wenyin Fu and
                  Brian Fuller and
                  Cynthia Gao and
                  Vedanuj Goswami and
                  Naman Goyal and
                  Anthony Hartshorn and
                  Saghar Hosseini and
                  Rui Hou and
                  Hakan Inan and
                  Marcin Kardas and
                  Viktor Kerkez and
                  Madian Khabsa and
                  Isabel Kloumann and
                  Artem Korenev and
                  Punit Singh Koura and
                  Marie{-}Anne Lachaux and
                  Thibaut Lavril and
                  Jenya Lee and
                  Diana Liskovich and
                  Yinghai Lu and
                  Yuning Mao and
                  Xavier Martinet and
                  Todor Mihaylov and
                  Pushkar Mishra and
                  Igor Molybog and
                  Yixin Nie and
                  Andrew Poulton and
                  Jeremy Reizenstein and
                  Rashi Rungta and
                  Kalyan Saladi and
                  Alan Schelten and
                  Ruan Silva and
                  Eric Michael Smith and
                  Ranjan Subramanian and
                  Xiaoqing Ellen Tan and
                  Binh Tang and
                  Ross Taylor and
                  Adina Williams and
                  Jian Xiang Kuan and
                  Puxin Xu and
                  Zheng Yan and
                  Iliyan Zarov and
                  Yuchen Zhang and
                  Angela Fan and
                  Melanie Kambadur and
                  Sharan Narang and
                  Aur{\'{e}}lien Rodriguez and
                  Robert Stojnic and
                  Sergey Edunov and
                  Thomas Scialom},
  title        = {Llama 2: Open Foundation and Fine-Tuned Chat Models},
  journal      = {CoRR},
  volume       = {abs/2307.09288},
  year         = {2023},
  url          = {https://doi.org/10.48550/arXiv.2307.09288},
  doi          = {10.48550/ARXIV.2307.09288},
  eprinttype    = {arXiv},
  eprint       = {2307.09288},
  bibsource    = {dblp computer science bibliography, https://dblp.org}
}

@article{DBLP:journals/corr/abs-2407-21783,
  author       = {Llama Team},
  title        = {The Llama 3 Herd of Models},
  journal      = {CoRR},
  volume       = {abs/2407.21783},
  year         = {2024},
  url          = {https://doi.org/10.48550/arXiv.2407.21783},
  doi          = {10.48550/ARXIV.2407.21783},
  eprinttype    = {arXiv},
  eprint       = {2407.21783},
  bibsource    = {dblp computer science bibliography, https://dblp.org}
}

@article{DBLP:journals/corr/abs-2401-02385,
  author       = {Peiyuan Zhang and
                  Guangtao Zeng and
                  Tianduo Wang and
                  Wei Lu},
  title        = {TinyLlama: An Open-Source Small Language Model},
  journal      = {CoRR},
  volume       = {abs/2401.02385},
  year         = {2024},
  url          = {https://doi.org/10.48550/arXiv.2401.02385},
  doi          = {10.48550/ARXIV.2401.02385},
  eprinttype    = {arXiv},
  eprint       = {2401.02385},
  bibsource    = {dblp computer science bibliography, https://dblp.org}
}

@article{DBLP:journals/corr/abs-2210-17323,
  author       = {Elias Frantar and
                  Saleh Ashkboos and
                  Torsten Hoefler and
                  Dan Alistarh},
  title        = {{GPTQ:} Accurate Post-Training Quantization for Generative Pre-trained
                  Transformers},
  journal      = {CoRR},
  volume       = {abs/2210.17323},
  year         = {2022},
  url          = {https://doi.org/10.48550/arXiv.2210.17323},
  doi          = {10.48550/ARXIV.2210.17323},
  eprinttype    = {arXiv},
  eprint       = {2210.17323},
  bibsource    = {dblp computer science bibliography, https://dblp.org}
}

@inproceedings{DBLP:conf/mlsys/0002TTYCWXDG024,
  author       = {Ji Lin and
                  Jiaming Tang and
                  Haotian Tang and
                  Shang Yang and
                  Wei{-}Ming Chen and
                  Wei{-}Chen Wang and
                  Guangxuan Xiao and
                  Xingyu Dang and
                  Chuang Gan and
                  Song Han},
  editor       = {Phillip B. Gibbons and
                  Gennady Pekhimenko and
                  Christopher De Sa},
  title        = {{AWQ:} Activation-aware Weight Quantization for On-Device {LLM} Compression
                  and Acceleration},
  booktitle    = {Proceedings of the Seventh Annual Conference on Machine Learning and
                  Systems, MLSys 2024, Santa Clara, CA, USA, May 13-16, 2024},
  publisher    = {mlsys.org},
  year         = {2024},
  url          = {https://proceedings.mlsys.org/paper\_files/paper/2024/hash/42a452cbafa9dd64e9ba4aa95cc1ef21-Abstract-Conference.html},
  bibsource    = {dblp computer science bibliography, https://dblp.org}
}

@inproceedings{DBLP:conf/acl/AlizadehMBKCMRF24,
  author       = {Keivan Alizadeh and
                  Iman Mirzadeh and
                  Dmitry Belenko and
                  S. Khatamifard and
                  Minsik Cho and
                  Carlo C. del Mundo and
                  Mohammad Rastegari and
                  Mehrdad Farajtabar},
  editor       = {Lun{-}Wei Ku and
                  Andre Martins and
                  Vivek Srikumar},
  title        = {{LLM} in a flash: Efficient Large Language Model Inference with Limited
                  Memory},
  booktitle    = {Proceedings of the 62nd Annual Meeting of the Association for Computational
                  Linguistics (Volume 1: Long Papers), {ACL} 2024, Bangkok, Thailand,
                  August 11-16, 2024},
  pages        = {12562--12584},
  publisher    = {Association for Computational Linguistics},
  year         = {2024},
  url          = {https://doi.org/10.18653/v1/2024.acl-long.678},
  doi          = {10.18653/V1/2024.ACL-LONG.678},
  bibsource    = {dblp computer science bibliography, https://dblp.org}
}

@article{DBLP:journals/csur/ZhengCQSSC25,
  author       = {Yue Zheng and
                  Yuhao Chen and
                  Bin Qian and
                  Xiufang Shi and
                  Yuanchao Shu and
                  Jiming Chen},
  title        = {A Review on Edge Large Language Models: Design, Execution, and Applications},
  journal      = {{ACM} Comput. Surv.},
  volume       = {57},
  number       = {8},
  pages        = {209:1--209:35},
  year         = {2025},
  url          = {https://doi.org/10.1145/3719664},
  doi          = {10.1145/3719664},
  bibsource    = {dblp computer science bibliography, https://dblp.org}
}

@inproceedings{DBLP:conf/iclr/HendrycksBBZMSS21,
  author       = {Dan Hendrycks and
                  Collin Burns and
                  Steven Basart and
                  Andy Zou and
                  Mantas Mazeika and
                  Dawn Song and
                  Jacob Steinhardt},
  title        = {Measuring Massive Multitask Language Understanding},
  booktitle    = {9th International Conference on Learning Representations, {ICLR} 2021,
                  Virtual Event, Austria, May 3-7, 2021},
  publisher    = {OpenReview.net},
  year         = {2021},
  url          = {https://openreview.net/forum?id=d7KBjmI3GmQ},
  bibsource    = {dblp computer science bibliography, https://dblp.org}
}

@article{DBLP:journals/tmlr/LiangBLTSYZNWKN23,
  author       = {Percy Liang and
                  Rishi Bommasani and
                  Tony Lee and
                  Dimitris Tsipras and
                  Dilara Soylu and
                  Michihiro Yasunaga and
                  Yian Zhang and
                  Deepak Narayanan and
                  Yuhuai Wu and
                  Ananya Kumar and
                  Benjamin Newman and
                  Binhang Yuan and
                  Bobby Yan and
                  Ce Zhang and
                  Christian Cosgrove and
                  Christopher D. Manning and
                  Christopher R{\'{e}} and
                  Diana Acosta{-}Navas and
                  Drew A. Hudson and
                  Eric Zelikman and
                  Esin Durmus and
                  Faisal Ladhak and
                  Frieda Rong and
                  Hongyu Ren and
                  Huaxiu Yao and
                  Jue Wang and
                  Keshav Santhanam and
                  Laurel J. Orr and
                  Lucia Zheng and
                  Mert Y{\"{u}}ksekg{\"{o}}n{\"{u}}l and
                  Mirac Suzgun and
                  Nathan Kim and
                  Neel Guha and
                  Niladri S. Chatterji and
                  Omar Khattab and
                  Peter Henderson and
                  Qian Huang and
                  Ryan Chi and
                  Sang Michael Xie and
                  Shibani Santurkar and
                  Surya Ganguli and
                  Tatsunori Hashimoto and
                  Thomas Icard and
                  Tianyi Zhang and
                  Vishrav Chaudhary and
                  William Wang and
                  Xuechen Li and
                  Yifan Mai and
                  Yuhui Zhang and
                  Yuta Koreeda},
  title        = {Holistic Evaluation of Language Models},
  journal      = {Trans. Mach. Learn. Res.},
  volume       = {2023},
  year         = {2023},
  url          = {https://openreview.net/forum?id=iO4LZibEqW},
  bibsource    = {dblp computer science bibliography, https://dblp.org}
}

@article{DBLP:journals/corr/abs-1911-02549,
  author       = {Vijay Janapa Reddi and
                  Christine Cheng and
                  David Kanter and
                  Peter Mattson and
                  Guenther Schmuelling and
                  Carole{-}Jean Wu and
                  Brian Anderson and
                  Maximilien Breughe and
                  Mark Charlebois and
                  William Chou and
                  Ramesh Chukka and
                  Cody Coleman and
                  Sam Davis and
                  Pan Deng and
                  Greg Diamos and
                  Jared Duke and
                  Dave Fick and
                  J. Scott Gardner and
                  Itay Hubara and
                  Sachin Idgunji and
                  Thomas B. Jablin and
                  Jeff Jiao and
                  Tom St. John and
                  Pankaj Kanwar and
                  David Lee and
                  Jeffery Liao and
                  Anton Lokhmotov and
                  Francisco Massa and
                  Peng Meng and
                  Paulius Micikevicius and
                  Colin Osborne and
                  Gennady Pekhimenko and
                  Arun Tejusve Raghunath Rajan and
                  Dilip Sequeira and
                  Ashish Sirasao and
                  Fei Sun and
                  Hanlin Tang and
                  Michael Thomson and
                  Frank Wei and
                  Ephrem Wu and
                  Lingjie Xu and
                  Koichi Yamada and
                  Bing Yu and
                  George Yuan and
                  Aaron Zhong and
                  Peizhao Zhang and
                  Yuchen Zhou},
  title        = {MLPerf Inference Benchmark},
  journal      = {CoRR},
  volume       = {abs/1911.02549},
  year         = {2019},
  url          = {http://arxiv.org/abs/1911.02549},
  eprinttype    = {arXiv},
  eprint       = {1911.02549},
  bibsource    = {dblp computer science bibliography, https://dblp.org}
}

@article{DBLP:journals/tmlr/ChenZ024,
  author       = {Lingjiao Chen and
                  Matei Zaharia and
                  James Zou},
  title        = {FrugalGPT: How to Use Large Language Models While Reducing Cost and
                  Improving Performance},
  journal      = {Trans. Mach. Learn. Res.},
  volume       = {2024},
  year         = {2024},
  url          = {https://openreview.net/forum?id=cSimKw5p6R},
  bibsource    = {dblp computer science bibliography, https://dblp.org}
}

@article{DBLP:journals/cacm/WilliamsWP09,
  author       = {Samuel Williams and
                  Andrew Waterman and
                  David A. Patterson},
  title        = {Roofline: an insightful visual performance model for multicore architectures},
  journal      = {Commun. {ACM}},
  volume       = {52},
  number       = {4},
  pages        = {65--76},
  year         = {2009},
  url          = {https://doi.org/10.1145/1498765.1498785},
  doi          = {10.1145/1498765.1498785},
  bibsource    = {dblp computer science bibliography, https://dblp.org}
}

@misc{jiang2025moecapbenchmarkingcostaccuracy,
      title={MoE-CAP: Benchmarking Cost, Accuracy and Performance of Sparse Mixture-of-Experts Systems}, 
      author={Yinsicheng Jiang and Yao Fu and Yeqi Huang and Ping Nie and Zhan Lu and Leyang Xue and Congjie He and Man-Kit Sit and Jilong Xue and Li Dong and Ziming Miao and Dayou Du and Tairan Xu and Kai Zou and Edoardo Ponti and Luo Mai},
      year={2025},
      eprint={2412.07067},
      archivePrefix={arXiv},
      primaryClass={cs.LG},
      url={https://arxiv.org/abs/2412.07067}, 
}

@article{DBLP:journals/corr/abs-2511-22334,
  author       = {Pablo Prieto and
                  Pablo Abad},
  title        = {Edge Deployment of Small Language Models, a comprehensive comparison
                  of CPU, {GPU} and {NPU} backends},
  journal      = {CoRR},
  volume       = {abs/2511.22334},
  year         = {2025},
  url          = {https://doi.org/10.48550/arXiv.2511.22334},
  doi          = {10.48550/ARXIV.2511.22334},
  eprinttype    = {arXiv},
  eprint       = {2511.22334},
  bibsource    = {dblp computer science bibliography, https://dblp.org}
}

@article{DBLP:journals/corr/abs-2511-05502,
  author       = {Varun Rajesh and
                  Om Jodhpurkar and
                  Pooja Anbuselvan and
                  Mantinder Singh and
                  Ashok Jallepali and
                  Shantanu Godbole and
                  Pradeep Kumar Sharma and
                  Hritvik Shrivastava},
  title        = {Production-Grade Local {LLM} Inference on Apple Silicon: {A} Comparative
                  Study of MLX, MLC-LLM, Ollama, llama.cpp, and PyTorch {MPS}},
  journal      = {CoRR},
  volume       = {abs/2511.05502},
  year         = {2025},
  url          = {https://doi.org/10.48550/arXiv.2511.05502},
  doi          = {10.48550/ARXIV.2511.05502},
  eprinttype    = {arXiv},
  eprint       = {2511.05502},
  bibsource    = {dblp computer science bibliography, https://dblp.org}
}
\bibliographystyle{icml2026}


\newpage
\appendix
\onecolumn

\section{Related Work} \label{sec:Appendix A Related Work}

\textbf{Trends in Efficient LLMs.} With the rapid proliferation of open-source
foundation models such as the Llama series
\cite{DBLP:journals/corr/abs-2307-09288} \cite{DBLP:journals/corr/abs-2407-21783} ,
scaling laws \cite{DBLP:journals/corr/abs-2001-08361} have revealed a positive
correlation between model capability and parameter scale, driving the trend toward
increasingly massive models. However, driven by concerns regarding data privacy,
latency sensitivity, and operational costs, deploying LLMs on on-device hardware
has become an imperative, catalyzing a fundamental paradigm shift from cloud-based
inference to edge computing \cite{DBLP:journals/csur/ZhengCQSSC25}. To accommodate constrained on-device resources, compact
architectures optimized for mobile scenarios have emerged, such as TinyLlama
\cite{DBLP:journals/corr/abs-2401-02385} and MobileLLM
\cite{DBLP:conf/icml/Liu0ILTFXCSKLC24} . Concurrently, mature post-training quantization
techniques like GPTQ \cite{DBLP:journals/corr/abs-2210-17323} and AWQ
\cite{DBLP:conf/mlsys/0002TTYCWXDG024} have significantly reduced memory footprints and computational
loads. Despite these advancements in model compression, as noted by Alizadeh et al.
\cite{DBLP:conf/acl/AlizadehMBKCMRF24} , the performance bottleneck for generative LLMs
on edge devices is typically strictly governed by memory bandwidth rather than
compute capacity (the "Memory Wall"). The limitations of these hardware characteristics necessitate a more profound, hardware-aware analysis of inference performance, moving beyond a mere evaluation of inference metrics.

\textbf{General Capabilities Benchmarking.}
Existing evaluation frameworks for Large Language Models (LLMs) predominantly focus on their cognitive and linguistic capabilities. Among these, MMLU \cite{DBLP:conf/iclr/HendrycksBBZMSS21} serves as a foundational benchmark, covering 57 tasks across diverse domains such as mathematics, history, and law to assess multi-task problem-solving abilities. Furthermore, HELM \cite{DBLP:journals/tmlr/LiangBLTSYZNWKN23}implements a holistic, multi-metric approach to evaluate dozens of mainstream LLMs, offering a comprehensive view of their semantic performance across various core scenarios.

While these benchmarks are essential for measuring "model intelligence," they typically treat the inference process as a system-agnostic black box. Consequently, they prioritize task-specific accuracy while disregarding the underlying computational overhead, such as system efficiency, real-time resource consumption, and hardware utilization. For resource-constrained edge devices, relying solely on these capability-centric metrics is insufficient to guarantee deployment viability, underscoring the urgent need for a more granular, system-level evaluation perspective.

\textbf{System Efficiency and On-Device Benchmarking.} To bridge the gap between algorithmic capability and physical deployment, recent research has shifted toward system-level efficiency. Industry standards like MLPerf \cite{DBLP:journals/corr/abs-1911-02549} establish rigorous protocols for evaluating inference systems across heterogeneous backends, while FrugalGPT \cite{DBLP:journals/tmlr/ChenZ024} introduces a cost-centric perspective, analyzing the trade-offs between API expenditures and model performance.

Specifically for on-device environments, several recent studies have provided in-depth empirical analyses. Edge Deployment of SLMs \cite{DBLP:journals/corr/abs-2511-22334} conducts a comprehensive evaluation across mobile CPUs, GPUs, and NPUs, highlighting the superior energy efficiency of NPUs and the necessity of bandwidth normalization for cross-architecture fairness. Within specific hardware ecosystems, Local LLM Inference on Apple Silicon \cite{DBLP:journals/corr/abs-2511-05502} systematically compares runtimes such as MLX and MLC-LLM, characterizing their throughput and latency within unified memory architectures. Furthermore, for sparse models, MoE-CAP \cite{jiang2025moecapbenchmarkingcostaccuracy} identifies the inherent trade-offs between Cost, Accuracy, and Performance (CAP), proposing sparsity-aware utilization metrics.

Despite these advancements, a fundamental limitation persists: existing benchmarks primarily report observed execution metrics (e.g., TTFT, throughput) without contextualizing them against the theoretical limits of the hardware. Such empirical observations often fail to decouple software-level optimizations from inherent hardware capabilities, potentially leading to biased conclusions regarding architectural efficiency. Our work addresses this gap by leveraging the Roofline Model \cite{DBLP:journals/cacm/WilliamsWP09}. By characterizing performance as a function of arithmetic intensity and hardware ceilings (peak compute and memory bandwidth), our framework enables a more transparent and fair benchmarking process that quantifies how effectively an LLM implementation approaches its hardware’s theoretical potential.

\section{Future Work} \label{sec:Appendix B Future Work}

While this study provides an extensive empirical characterization of mainstream small language models across diverse hardware tiers, we identify several strategic avenues for future exploration. Primarily, we intend to expand the taxonomy of our benchmarking framework to incorporate a broader range of attention mechanisms. Although our current analysis provides a comprehensive evaluation of critical patterns such as $MLA$ and $GQA$, we aim to adapt our methodology to emerging structural variants and hybrid architectures to ensure the continued relevance of our operational intensity ($OI$) insights across the evolving architectural landscape.

A significant frontier lies in the analytical characterization of Mixture of Experts (MoE) architectures. Quantifying the theoretical $FLOPS$ for MoE models during dynamic inference remains a formidable challenge because the stochastic nature of token routing complicates the estimation of the execution ceiling. While the current work focuses on dense architectures, extending the Roofline model to account for sparse activation regimes will be essential for identifying the unique physical constraints and efficiency traps inherent in localized MoE deployment.

Finally, we plan to broaden the experimental scope across both hardware and software dimensions. On the hardware side, we aim to scale our testing to a more diverse array of heterogeneous edge devices to further validate the cross platform robustness of our findings. On the software side, we recognize that different inference engines, such as TensorRT LLM, vLLM, and ONNX Runtime, introduce substantial variance in performance. Investigating how these software stacks influence $OI$ and realized throughput will provide a more holistic understanding of the entire deployment pipeline, facilitating a more effective transition toward true hardware software co design.

\section{Experimental Details}\label{sec:Appendix C Experimental Details}
The  code for this work is available at \url{https://github.com/banbu-ai/roofline_bench}.

\subsection{Evaluated Model Architectures}

To investigate the scaling laws and efficiency of intelligence on-device, we select a diverse set of models that represent current trends in localized AI. These models are evaluated across two categories based on the computational tiers of the hardware substrates:

\begin{itemize}
\item \textbf{Edge-Scale Models}: For the most resource-limited tier represented by the Raspberry Pi 5, we evaluate ultra-compact models including the Pythia (160M, 410M), Qwen2.5-0.5B, and SmolLM2 (135M, 360M) families. These models primarily utilize Q8\_0 quantization to fit within the constraints of general-purpose CPUs.
\item \textbf{Mobile-Class Models}: For performance-oriented platforms such as the Apple M1 Pro, RTX 3070 Ti Laptop, and Jetson Orin Nano Super, we employ models ranging from 0.6B to 1.8B parameters. This group features diverse types of attention to observe their impact on the potential for inference, including Grouped-Query Attention (GQA) for Qwen2.5-1.5B, Llama-3.2-1B, Qwen3-0.6B, and Fox-1-1.6B; Multi-Head Attention (MHA) for SmolLM2-1.7B; and Multi-Head Latent Attention (MLA) for PLM-1.8B.
\end{itemize}

All architectures are tested under four distinct scenarios—SISO, LISO, SILO, and LILO—to capture performance dynamics across varying sequence lengths. This selection provides a rigorous basis for the co-design of models and hardware by identifying how structural primitives like attention mechanisms interact with physical memory boundaries.

\subsection{Evaluated Hardware Substrates}

To ensure a rigorous evaluation across various tiers of computation, we conduct our analysis on four distinct hardware platforms. These substrates represent a broad spectrum of architectural designs and power envelopes, reflecting the diverse physical constraints encountered in on-device intelligence. The selection includes the Apple M1 Pro (a unified system on a chip), the RTX 3070 Ti Laptop (a discrete GPU for mobile workstations), the Jetson Orin Nano Super (a dedicated module for AI at the edge), and the Raspberry Pi 5 (a ubiquitous platform based on a general CPU).

Table~\ref{tab:hardware_specs} summarizes the specifications for heterogeneous platforms, ranging from Discrete GPUs to General CPUs. By incorporating theoretical and measured values for peak performance ($\pi$) and memory bandwidth ($\beta$), these metrics establish the boundaries for calculating Relative Inference Potential ($\Phi$). Such characterization is essential to identify requirements for effective hardware-software co-design in intelligence on-device.


\begin{table*}[ht]
\caption{Hardware Specifications and Performance Metrics across Heterogeneous Platforms. Theo denotes the theoretical peak value provided by the manufacturer, while Meas denotes the measured peak performance under benchmarking workloads. Notably, the measured FP16 peak performance for NVIDIA RTX 30 series GPUs and Jetson Orin Nano significantly exceeds their theoretical FP32 peak values because of the activation of specialized hardware such as Tensor Cores during evaluation. Values marked with $\dagger$ are in GFLOPS, while all other compute metrics are in TFLOPS.}
\label{tab:hardware_specs}
\begin{center}
\begin{small}
\begin{tabular}{lcccc}
\toprule
\textbf{Chip} & \textbf{Architecture} & \textbf{Bandwidth (GB/s)} & \textbf{FP16 Peak (TFLOPS)} & \textbf{FP32 Peak (TFLOPS)} \\
 & & \textbf{Theo / Meas} & \textbf{Theo / Meas} & \textbf{Theo / Meas} \\
\midrule
NVIDIA RTX 3090           & Discrete GPU    & 936.20 / 560.02 & 35.58 / 66.20        & 35.58 / 24.28 \\
NVIDIA RTX 3070 Ti Laptop & Discrete GPU    & 448.00 / 217.00 & 16.60 / 31.76        & 16.60 / 9.51 \\
Apple M1 Pro              & Unified SoC     & 204.80 / 120.03 & --- / 4.61           & 5.20 / 4.31 \\
Jetson Orin Nano Super 8G & Edge AI Module  & 102.00 / 59.40  & 4.178 / 9.56         & 2.089 / 1.34 \\
Raspberry Pi 5            & General CPU     & 17.10 / 3.98    & --- / 1.48$^\dagger$ & 153.60$^\dagger$ / 78.56$^\dagger$ \\
\bottomrule
\end{tabular}
\end{small}
\end{center}
\end{table*}

\subsection{Experimental Setting}

\textbf{Hardware Characterization and Roofline Benchmarking.} To establish the empirical bounds for the Roofline model, we evaluate the peak throughput and memory bandwidth across a heterogeneous suite of devices, including macOS, Windows, NVIDIA Jetson Orin Nano, and Raspberry Pi platforms. We conduct synthetic benchmarks involving large-scale matrix addition and multiplication using the PyTorch framework. To ensure optimal performance across architectures, we utilize the CUDA backend for NVIDIA hardware and the Metal Performance Shaders (MPS) backend for Apple Silicon devices. These measurements provide the hardware-specific constants required for our performance analysis.

\textbf{FLOPs Calculation Methodology.} 
To evaluate operational intensity ($OI$) and performance within the Roofline model, we estimate the floating point operations ($FLOPs$) associated with each inference configuration. This estimation framework accounts for linear projections, attention mechanisms, and KV cache management by leveraging key architectural parameters including the hidden dimension ($h$), the number of query heads ($n_h$), and the number of key-value heads ($n_{kv}$). The specific estimation logic \citet{DBLP:journals/corr/abs-2506-17286}, with detailed formulas provided in Table~\ref{table:cal_flops}. The resulting $FLOPs$ values serve as an approximated baseline for deriving $OI$ and throughput metrics.

\begin{table*}[h]
\centering
\caption{Comparison of computational complexity and memory requirements for different attention mechanisms. $H$ is the hidden dimension, $N$ is the sequence length, $n_q, n_k, n_v, n_c$ are the number of query, key, value, and latent value heads, respectively; $d_h$ is the per-head dimension, $d_l$ is the latent dimension. For MLA-based structures, $d_c$ denotes the KV compression dimension, while $d_{rope}$ and $d_{nope}$ represent the rotary and non-rotary components of the embedding dimension, respectively.}
\resizebox{1.0\textwidth}{!}{
\begin{tabular}{lcccc}
\toprule
\multicolumn{1}{c}{\multirow{2}{*}{Attention}} & \multirow{2}{*}{KV Cache per Layer} & \multicolumn{2}{c}{Computation per Layer}                                                  & \multirow{2}{*}{Expressivity} \\
\multicolumn{1}{c}{}                                       &                                     & Attention                 & Linear                                                         &                             \\ \midrule
\textbf{MHA}                                 & $2n_hd_hN$                               & $2n_hd_hN^2$                    & $4NH^2$                                                         & Strong                      \\
\textbf{GQA}                              & $2n_kd_hN$                       & $2n_hd_hN^2$                    & $2NH^2+2n_kd_hNH$                                                & Moderate                    \\
\textbf{MLA}                         & $(d_c+d_{rope})N$                   & $n_h(d_{rope}+2d_{nope})N^2$ & $\Big((d_c+d_{rope})H+n_h(d_{rope}+d_{nope})H+2n_hd_ld_{nope}+H^2\Big)N$ & Strong                    \\
\textbf{GVA}                              & $(H+n_kd_h)N$                       & $(n_qd_h+n_hd_h)N^2$               & $2NH^2+2n_kd_hNH$                                                & Moderate                    \\
\textbf{GHA}                               & $(n_kd_h + n_vd_h)N$           & $(n_qd_h+n_hd_h)N^2$          & $NH^2+n_qd_hNH+n_kd_hNH+n_vd_hNH$                             & Weak                        \\
\textbf{GTA}                        & {$(n_kd_h + n_cd_l)N$}           & {$n_q(d_k+d_l)N^2$}          & {$2NH^2+(n_qd_h+n_kd_h+n_cd_l+d_l)NH$}                             & Strong                    \\ \bottomrule
\end{tabular}%
}
\label{table:cal_flops}
\end{table*}

\textbf{Inference Memory Monitoring.} We employ platform-specific strategies to monitor memory utilization during llama-bench execution. For devices with Unified Memory Architectures (UMA)—including macOS and NVIDIA Jetson—as well as CPU-based platforms like the Raspberry Pi 5, we track physical memory usage by mapping the target process name to its Process ID (PID) and sampling its resident set size (RSS) during inference.
On Windows systems equipped with consumer-grade discrete GPUs, monitoring process-specific VRAM presents a technical challenge. Due to the Windows Display Driver Model (WDDM) constraints on consumer hardware—which lack support for the Tesla Compute Cluster (TCC) mode required for granular per-process accounting—standard utilities such as nvidia-smi and pynvml often return null values for individual process memory. To circumvent this, we implement an exclusive-access monitoring strategy: by parsing nvidia-smi XML outputs, we ensure the discrete GPU is dedicated solely to the target PID during benchmarking. Under these isolated conditions, the total VRAM utilization of the device serves as a precise proxy for the inference memory footprint.

\textbf{Inference Workflow and Data Acquisition.} For each experimental trial, we execute llama-bench with parameterized input/output sequence lengths and thread counts. To ensure temporal synchronization across data streams, each session is assigned a unique timestamp. During execution, our framework concurrently monitors the memory footprint and records the runtime metrics. The resulting inference logs, formatted in JSON as per llama-bench specifications, capture comprehensive metadata including device architecture, backend configurations, model types, and specific command-line arguments. Upon completion of each inference cycle, our framework performs automated post-hoc analysis and computational evaluation based on these synchronized logs and memory traces.

\textbf{Memory Management and Offloading Strategy.} While the inference framework allows for explicit control over offloading, we incorporate this strategy to handle scenarios where model depth exceeds the 8 GB VRAM capacity of the RTX 3070Ti Laptop. The observed linear performance degradation in specific experimental trials is primarily attributed to the automated offloading mechanism in llama.cpp. When the GPU-resident buffers are insufficient to accommodate weight tensors, the framework dynamically reallocates storage to system memory. This allocation follows a tiered priority hierarchy: ACCEL (accelerated CPU regions) $\rightarrow$ GPU Host (pinned memory for optimized DMA transfer) $\rightarrow$ CPU Extra $\rightarrow$ Standard CPU (general-purpose system memory).

\section{Comprehensive Benchmarking Analysis of Model Architectures on Heterogeneous Hardware Platforms}\label{sec:Appendix D Benchmark Results}

This appendix provides a comprehensive suite of benchmarking results to substantiate the empirical findings and theoretical analyses presented in the main text. We detail the performance profiles of various large language model architectures including Multi head Latent Attention and trainable sparse attention across a spectrum of heterogeneous hardware platforms ranging from high performance accelerators to edge side SoCs. The presented data encompass a wide array of quantization precisions and operational scenarios while offering granular insights into the relationship between architectural design, layer wise operational intensity ($OI$), and hardware resource saturation. By providing these extended datasets, we aim to offer a rigorous validation of the scaling dynamics and utilization equity discussed throughout the study.

\subsection{Analysis and Findings on the Optimality Gap}

\begin{figure*}[hb]
    \centering
    \includegraphics[width=1.0\textwidth]{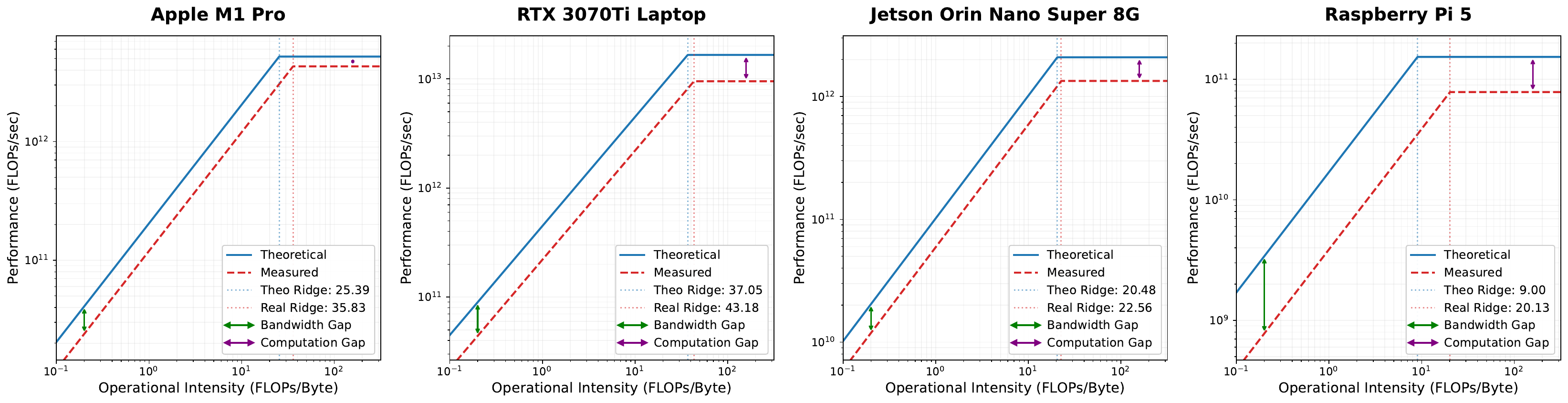}
    \caption{
    Performance gap analysis using Roofline models on different hardware. Green and purple arrows represent Bandwidth and Compute Gaps between theoretical (solid blue) and measured (dashed red) performance. Vertical lines mark the theoretical and real-world ridge points.}
    \label{fig:multi_device_roofline_gap}
\end{figure*}

The vertical distance between the actual operating point and the theoretical Roofline ceiling in the performance chart (as illustrated across different devices in Figure \ref{fig:multi_device_roofline_gap}). This gap quantifies the "left-on-the-table" performance and serves as a direct indicator of optimization headroom.

\subsection{Extended Benchmark Results: Prefilling and Decoding TPS}

\begin{table}[ht]
\caption{Detailed inference throughput (TPS) on Raspberry Pi 5 across various scenarios. Data are presented as Prefilling TPS / Decoding TPS.}
\label{tab:raspberry_pi5_tps}
\centering
\begin{small}
\begin{tabular}{l l c c c c}
\toprule
\textbf{Model} & \textbf{Precision} & \textbf{SISO} & \textbf{LISO} & \textbf{SILO} & \textbf{LILO} \\
\midrule
pythia-160m & Q8\_0 & 149 / 70 & 63 / 71 & 150 / 25 & 64 / 27 \\
pythia-410m & Q8\_0 & 43 / 26 & 20 / 26 & 43 / 9 & 21 / 9 \\
Qwen2.5-0.5B-Instruct & Q8\_0 & 38 / 20 & 22 / 20 & 38 / 14 & 22 / 14 \\
SmolLM2-135M-Instruct & Q8\_0 & 115 / 65 & 42 / 65 & 115 / 29 & 42 / 29 \\
SmolLM2-360M-Instruct & Q8\_0 & 41 / 26 & 20 / 26 & 42 / 14 & 20 / 14 \\
\bottomrule
\end{tabular}
\end{small}
\vskip 0.1in
\end{table}

\begin{table}[hb]
\caption{Cross device inference throughput (TPS) comparison for Qwen2.5-1.5B-Instruct in Q8\_0 precision. Data are presented as Prefilling TPS / Decoding TPS.}
\label{tab:qwen_cross_device_tps}
\centering
\begin{small}
\begin{tabular}{l l l c c c c}
\toprule
\textbf{Model} & \textbf{Precision} & \textbf{Device} & \textbf{SISO} & \textbf{LISO} & \textbf{SILO} & \textbf{LILO} \\
\midrule
\multirow{3}{*}{Qwen2.5-1.5B-Instruct} & \multirow{3}{*}{Q8\_0} & M1Pro & 1349 / 63 & 1204 / 62 & 1349 / 52 & 1204 / 52 \\
 &  & RTX 3070Ti L & 4465 / 127 & 5286 / 130 & 4295 / 118 & 5256 / 119 \\
 &  & Jetson Orin & 1203 / 25 & 1075 / 25 & 1281 / 23 & 1073 / 23 \\
\bottomrule
\end{tabular}
\end{small}
\end{table}

\begin{table}[ht]
\caption{Detailed inference throughput (TPS) on Apple M1 Pro across various scenarios. Data are presented as Prefilling TPS / Decoding TPS.}
\label{tab:m1pro_tps_full}
\centering
\begin{small}
\begin{tabular}{l l c c c c}
\toprule
\textbf{Model} & \textbf{Precision} & \textbf{SISO} & \textbf{LISO} & \textbf{SILO} & \textbf{LILO} \\
\midrule
\multirow{3}{*}{Qwen2.5-1.5B-Instruct} & FP16 & 1424 / 50 & 1273 / 50 & 1423 / 43 & 1273 / 43 \\
 & Q8\_0 & 1349 / 63 & 1204 / 62 & 1349 / 52 & 1204 / 52 \\
 & Q4\_K\_M & 1199 / 90 & 1078 / 90 & 1200 / 69 & 1078 / 69 \\
\midrule
\multirow{3}{*}{Qwen3-0.6B} & FP16 & 3168 / 102 & 2140 / 102 & 3164 / 69 & 2143 / 71 \\
 & Q8\_0 & 2984 / 105 & 2072 / 105 & 3018 / 72 & 2070 / 72 \\
 & Q4\_K\_M & 2765 / 100 & 1939 / 160 & 2770 / 94 & 1942 / 94 \\
\midrule
\multirow{3}{*}{Llama-3.2-1B-Instruct} & FP16 & 2002 / 65 & 1594 / 65 & 2000 / 51 & 1594 / 51 \\
 & Q8\_0 & 1894 / 92 & 1509 / 92 & 1894 / 67 & 1509 / 67 \\
 & Q4\_K\_M & 1691 / 133 & 1369 / 133 & 1691 / 86 & 1370 / 85 \\
\midrule
\multirow{3}{*}{PLM-1.8B-Instruct} & FP16 & 1114 / 40 & 939 / 40 & 1113 / 30 & 939 / 30 \\
 & Q8\_0 & 1053 / 55 & 891 / 55 & 1053 / 38 & 890 / 38 \\
 & Q4\_K\_M & 942 / 75 & 812 / 75 & 942 / 46 & 812 / 46 \\
\midrule
\multirow{3}{*}{Fox-1-1.6B} & FP16 & 1522 / 47 & 1239 / 47 & 1521 / 38 & 1239 / 38 \\
 & Q8\_0 & 1438 / 63 & 1174 / 63 & 1438 / 49 & 1174 / 49 \\
 & Q4\_K\_M & 1295 / 87 & 1070 / 88 & 1295 / 62 & 1071 / 62 \\
\midrule
\multirow{3}{*}{SmolLM2-1.7B-Instruct} & FP16 & 1212 / 46 & 970 / 46 & 1212 / 35 & 971 / 35 \\
 & Q8\_0 & 1139 / 64 & 918 / 64 & 1138 / 46 & 916 / 46 \\
 & Q4\_K\_M & 1002 / 90 & 827 / 90 & 1002 / 58 & 827 / 58 \\
\bottomrule
\end{tabular}
\end{small}
\end{table}

\clearpage

\subsection{Extended Empirical Characterization: Comprehensive Roofline Profiles}

\begin{figure*}[htbp]
    \centering
    \includegraphics[width=0.75\textwidth]{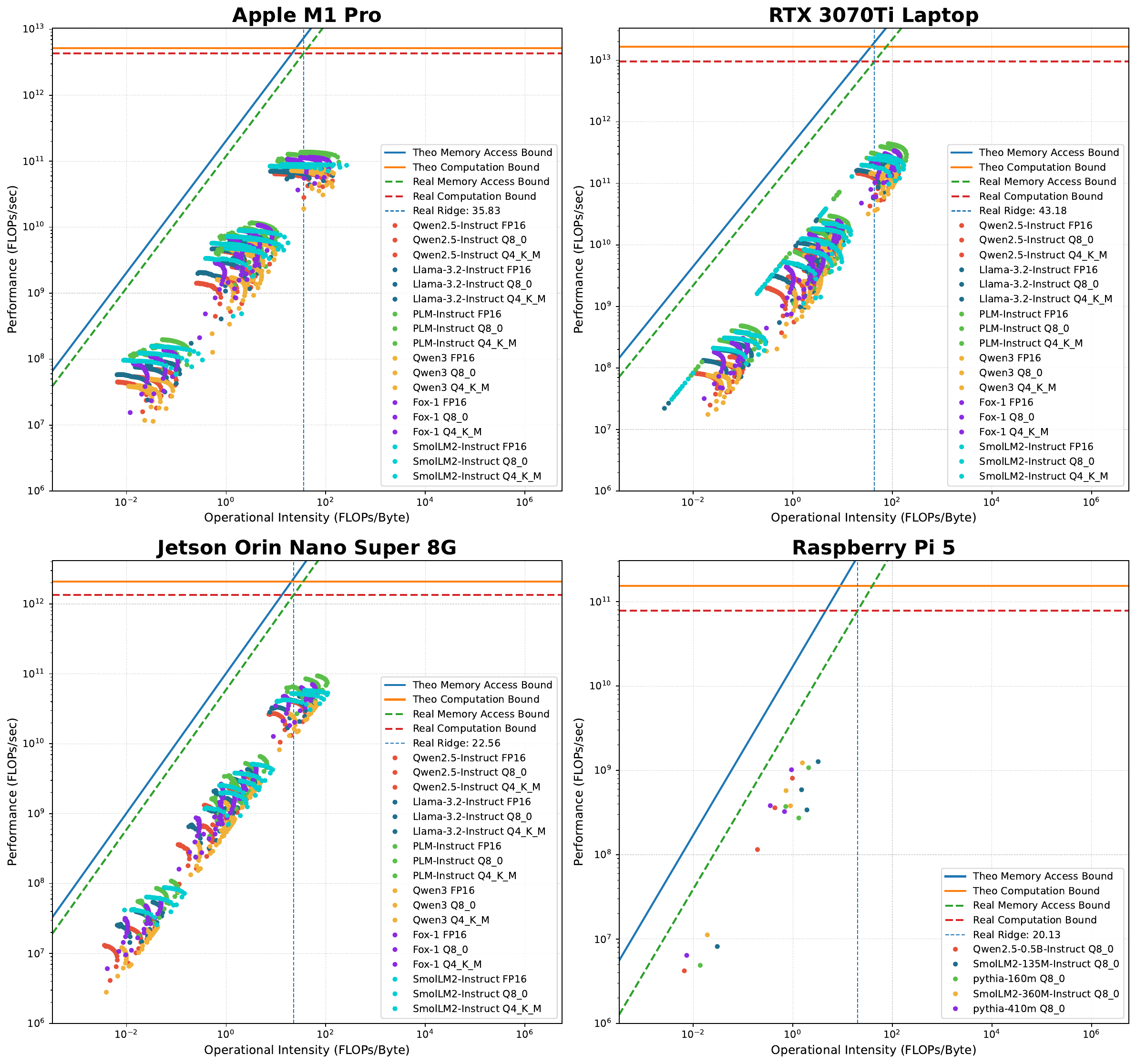}
    \caption{Hardware Specifications and Performance Metrics across Heterogeneous Platforms. Theo denotes the theoretical peak value provided by the manufacturer, while Meas denotes the measured peak performance under benchmarking workloads. Notably, the measured FP16 peak performance for NVIDIA RTX 30 series GPUs and Jetson Orin Nano significantly exceeds their theoretical FP32 peak values because of the activation of specialized hardware such as Tensor Cores during evaluation. Values marked with $\dagger$ are in GFLOPS, while all other compute metrics are in TFLOPS.}
    \label{fig:Overall_Benchmark_Analysis}
\end{figure*}


\begin{figure*}[htbp]
    \centering
    \includegraphics[width=1.0\textwidth]{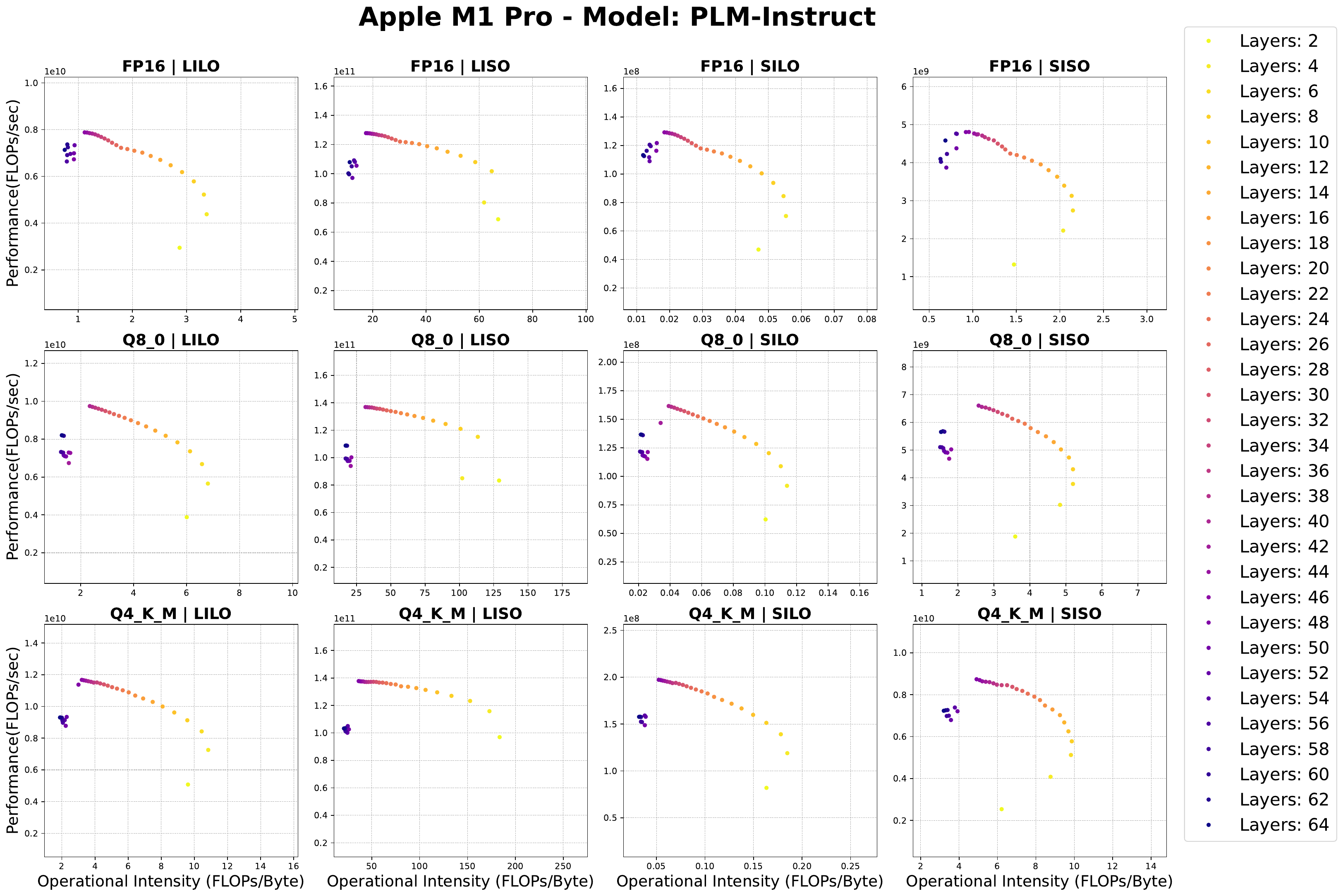}
    \caption{Layer-wise operational intensity trends for PLM-Instruct on Apple M1 Pro. This evaluation monitors the progression of computational intensity throughout the layer stack under diverse inference workloads. The profiles reveal that while most layers maintain a stable intensity, certain points exhibit a steep drop in metrics. Such abrupt decreases indicate that the model has reached a performance bottleneck, where the throughput is increasingly restricted by the underlying hardware architecture, leading to a noticeable reduction in layer-wise efficiency.}
\end{figure*}

\begin{figure*}[htbp]
    \centering
    \includegraphics[width=1.0\textwidth]{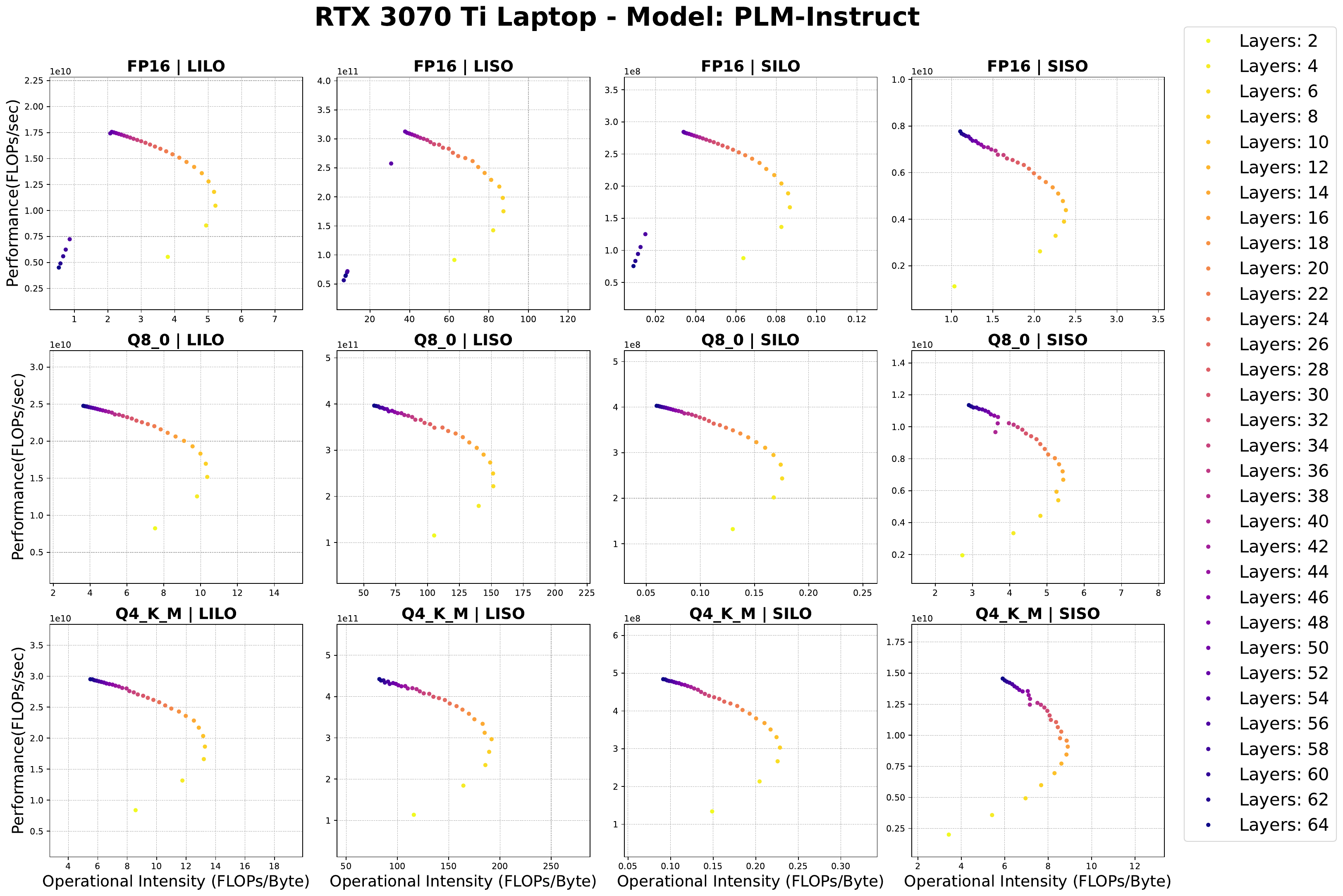}
    \caption{Layer-wise operational intensity trends for PLM-Instruct on NVIDIA RTX 3070 Ti Laptop. The subplots map the operational intensity (FLOPs/Byte) against the Transformer layer index across diverse inference workloads. Although the computational intensity is generally uniform throughout the layer stack, certain data points exhibit a significant downward trend. Such abrupt decreases mark the onset of hardware-specific bottlenecks, where the data movement or processing constraints lead to a noticeable reduction in layer-wise efficiency compared to the stable segments of the model.}
\end{figure*}

\begin{figure*}[htbp]
    \centering
    \includegraphics[width=1.0\textwidth]{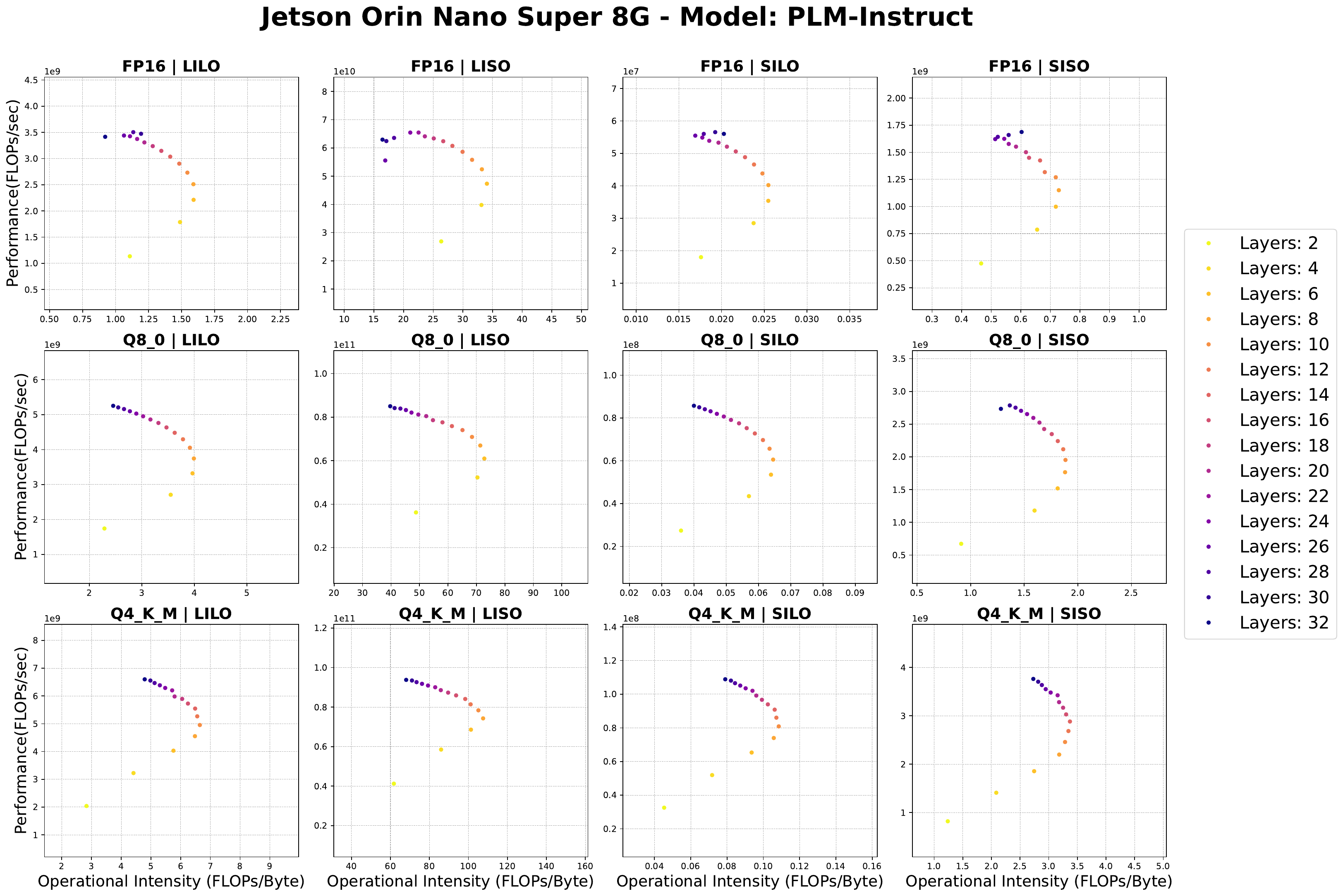}
    \caption{Layer-wise operational intensity trends for PLM-Instruct on Jetson Orin Nano Super 8G. This figure evaluates the distribution of computational intensity across Transformer layers for FP16, Q8\_0, and Q4\_K\_M precisions. While the operational intensity typically maintains a steady profile throughout the model depth, the sharp declines observed in specific scenarios signify that the execution has encountered critical hardware bottlenecks. These straight-line drops indicate regions where the performance is heavily constrained by system resource limits, preventing the layers from sustaining their peak arithmetic intensity.}
\end{figure*}

\begin{figure*}[htbp]
    \centering
    \includegraphics[width=1.0\textwidth]{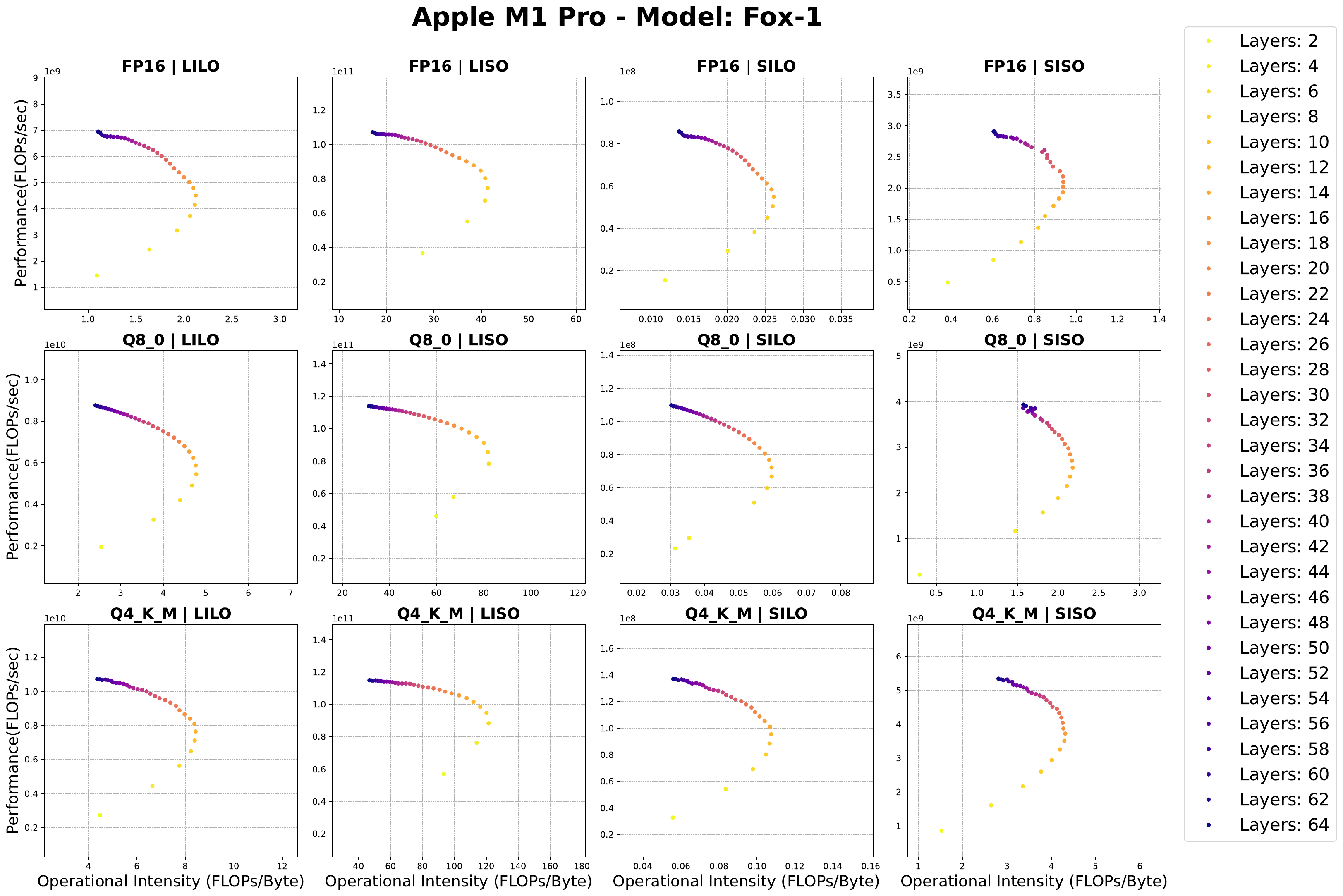}
    \caption{Layer-wise operational intensity trends for Fox-1 on Apple M1 Pro. This figure analyzes the stability of computational intensity across successive Transformer layers under FP16, Q8\_0, and Q4\_K\_M precisions. Across the diverse inference scenarios (SISO, SILO, LISO, and LILO), the operational intensity remains remarkably constant as the layer index increases. This horizontal trend suggests that the ratio of arithmetic operations to memory traffic is intrinsically determined by the model architecture and scenario type rather than the specific depth of the layer.}
\end{figure*}

\begin{figure*}[htbp]
    \centering
    \includegraphics[width=1.0\textwidth]{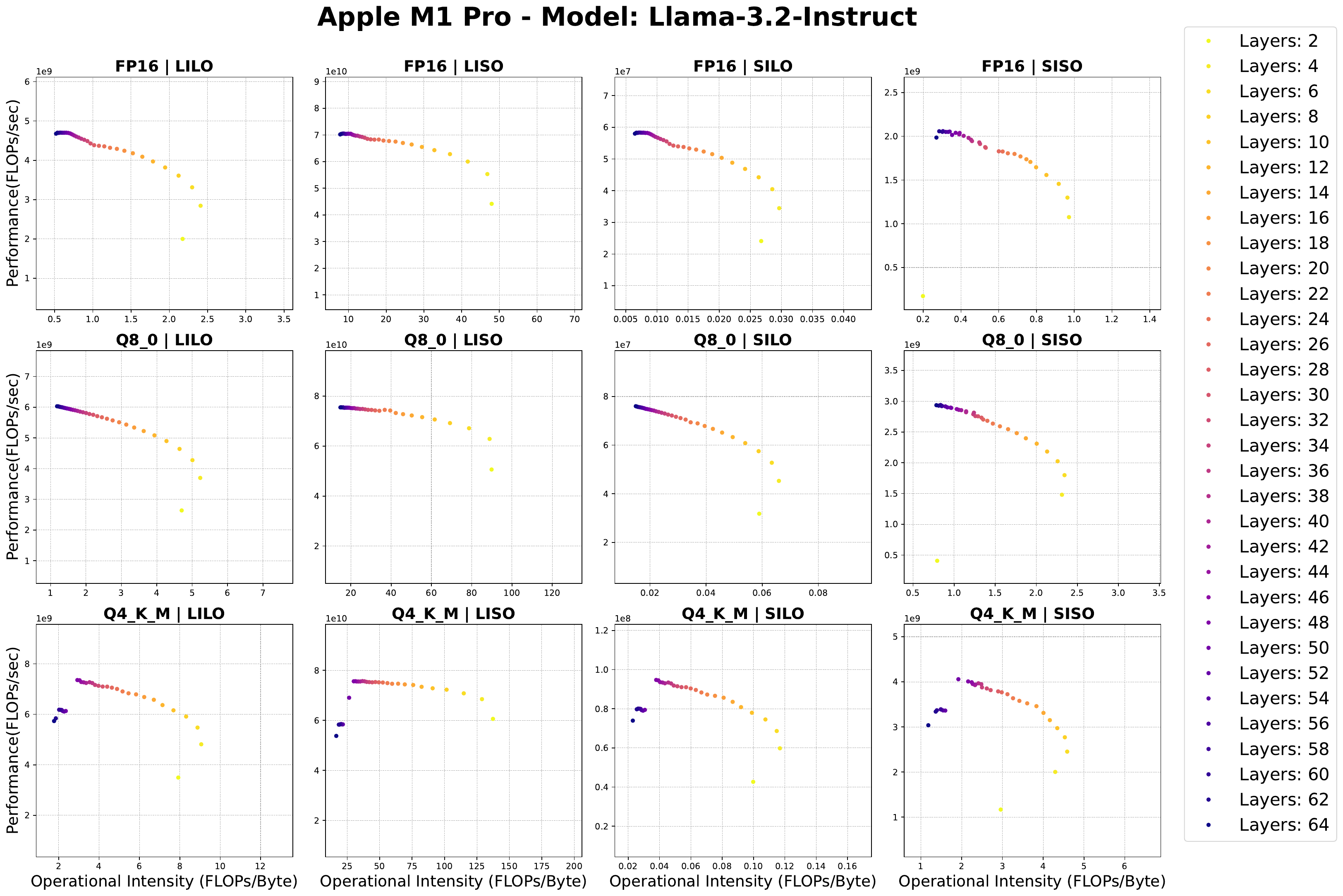}
    \caption{Layer-wise operational intensity trends for Llama-3.2-Instruct on Apple M1 Pro. The subplots illustrate the distribution of FLOPs/Byte across Transformer layers for FP16, Q8\_0, and Q4\_K\_M precisions. While the operational intensity is generally consistent across the model depth, a sharp decline is observed in specific scenarios, signifying that the execution has encountered a critical hardware bottleneck. This transition highlights the segments where the computational efficiency is significantly hampered by data movement constraints or other resource limitations.}
\end{figure*}

\begin{figure*}[htbp]
    \centering
    \includegraphics[width=1.0\textwidth]{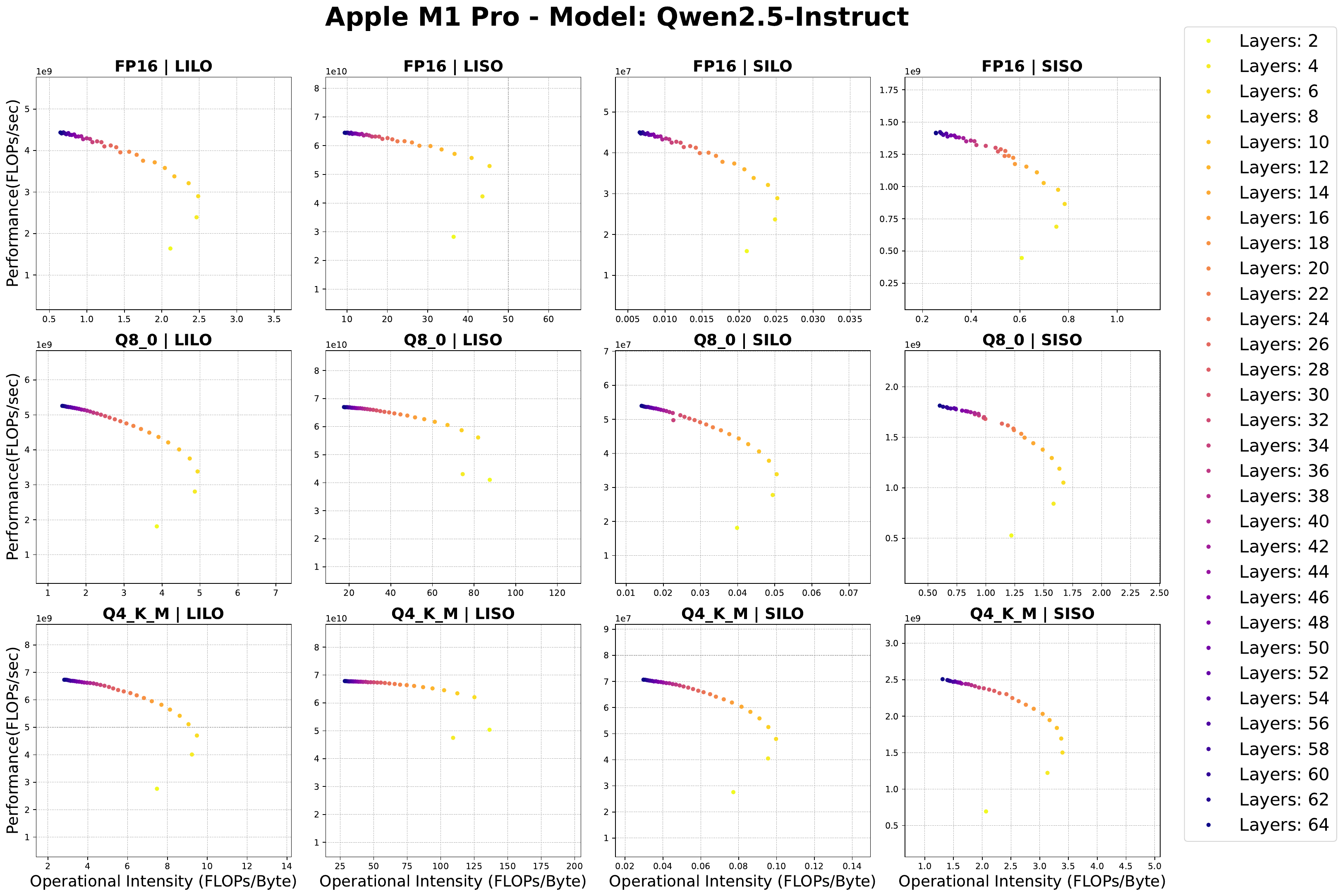}
    \caption{Layer-wise operational intensity trends for Qwen2.5-Instruct on Apple M1 Pro. These profiles characterize the layer-wise behavior of arithmetic intensity across multiple inference scenarios. The horizontal alignment of data points across the layer index confirms that the computational intensity is invariant to layer depth. This stability is maintained across all tested precisions, demonstrating that quantization shifts the absolute intensity but does not alter the uniform trend across the model's structural hierarchy.}
\end{figure*}

\begin{figure*}[htbp]
    \centering
    \includegraphics[width=1.0\textwidth]{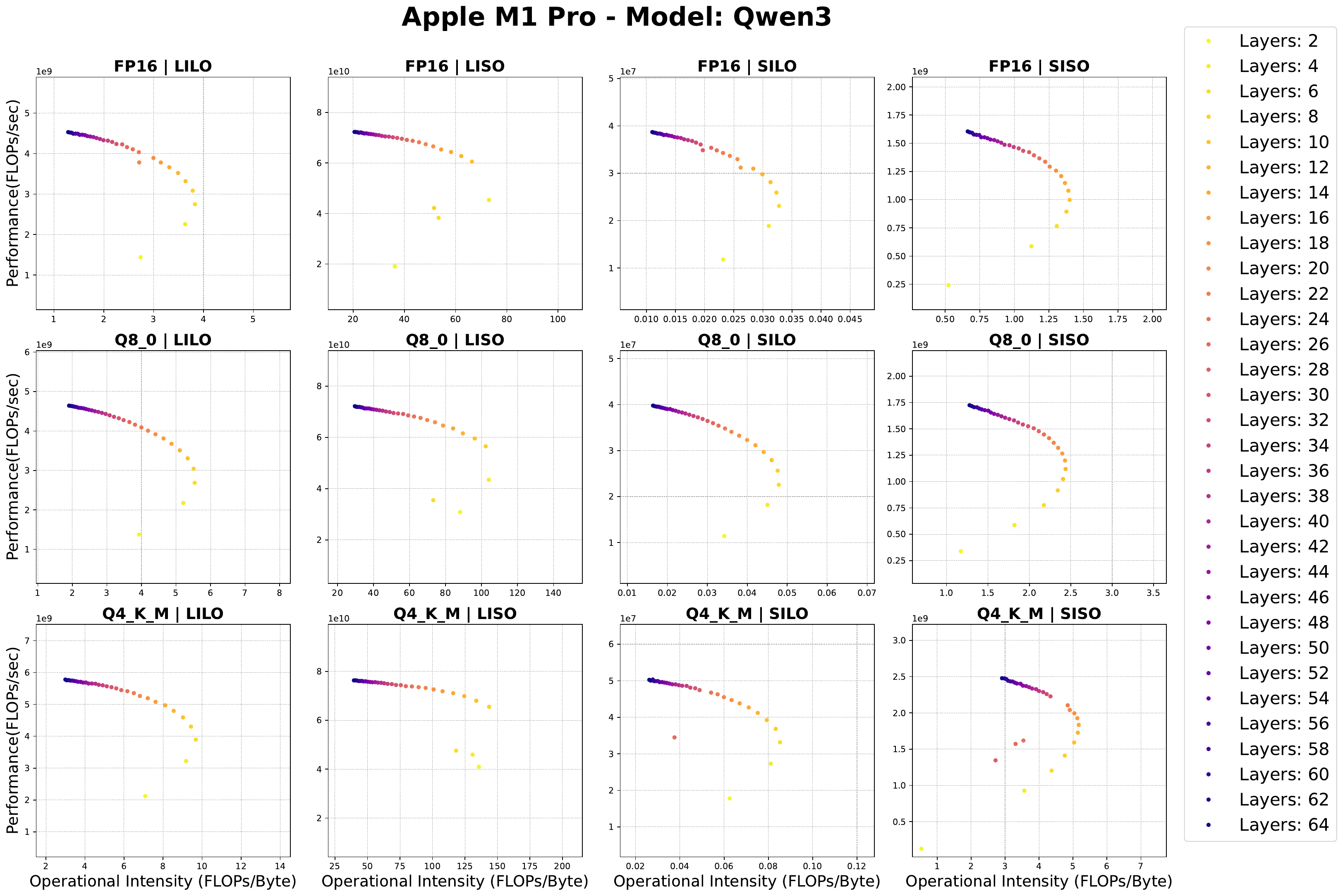}
    \caption{Layer-wise operational intensity trends for Qwen3 on Apple M1 Pro. The subplots map the operational intensity against the Transformer layer index for various precisions. Regardless of the inference workload (SISO, SILO, LISO, or LILO), the metrics exhibit a flat progression from the initial to the final layers. This trend indicates that the bottleneck characteristics, whether memory-bound or compute-bound, are shared equally by all layers within the model architecture.}
\end{figure*}


\begin{figure*}[htbp]
    \centering
    \includegraphics[width=0.75\textwidth]{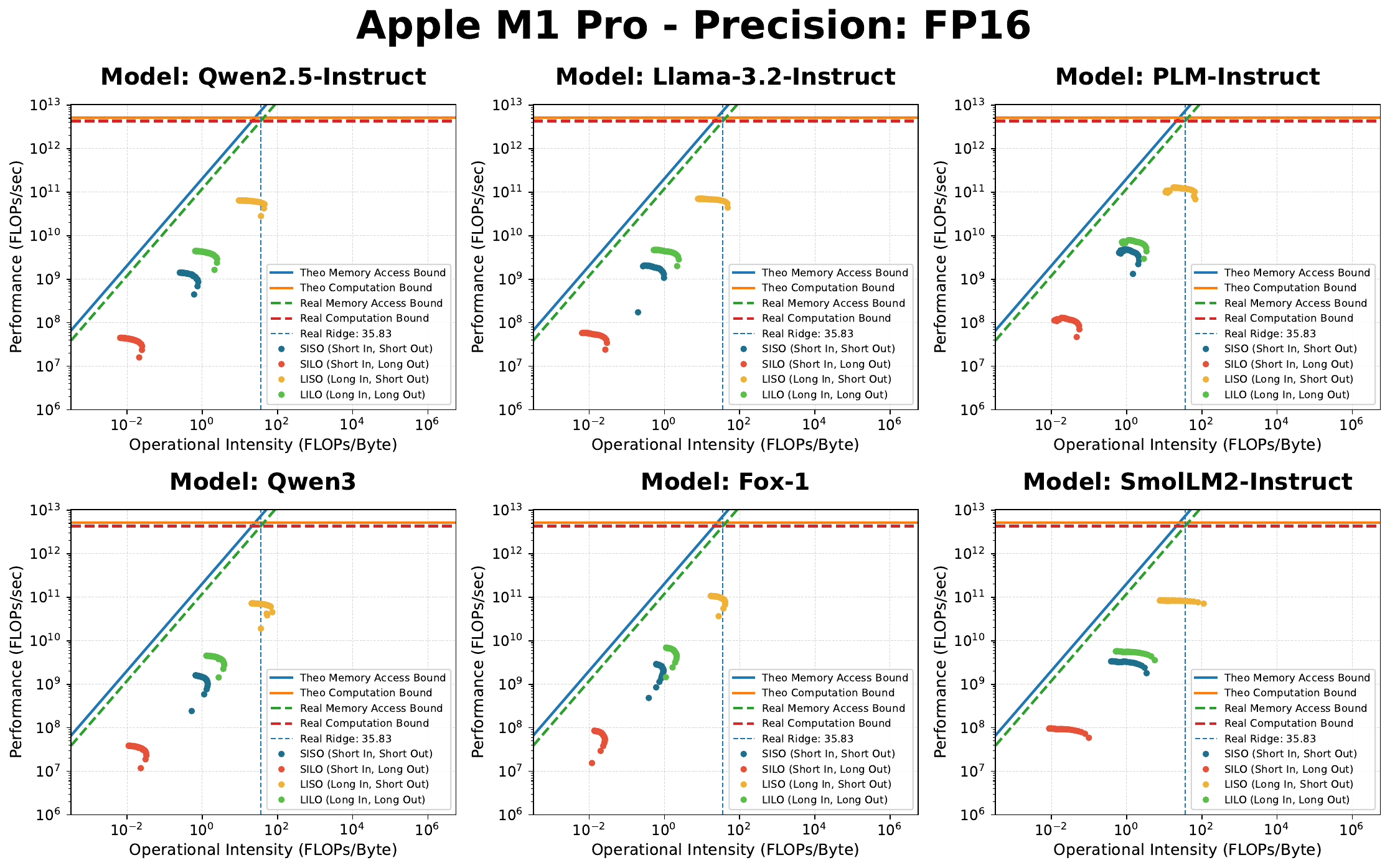}
    \caption{Performance profiles of diverse inference scenarios on Apple M1 Pro (FP16). This figure evaluates various models across four distinct operational scenarios: SISO (Short In, Short Out), SILO (Short In, Long Out), LISO (Long In, Short Out), and LILO (Long In, Long Out). The results illustrate how operational intensity (FLOPs/Byte) shifts performance (FLOPs/sec). LISO scenarios typically reach the real computation bound , while decoding-heavy SILO/LILO scenarios remain memory-bound. The hardware exhibits a real ridge point of 35.83.}
\end{figure*}

\begin{figure*}[htbp]
    \centering
    \includegraphics[width=0.75\textwidth]{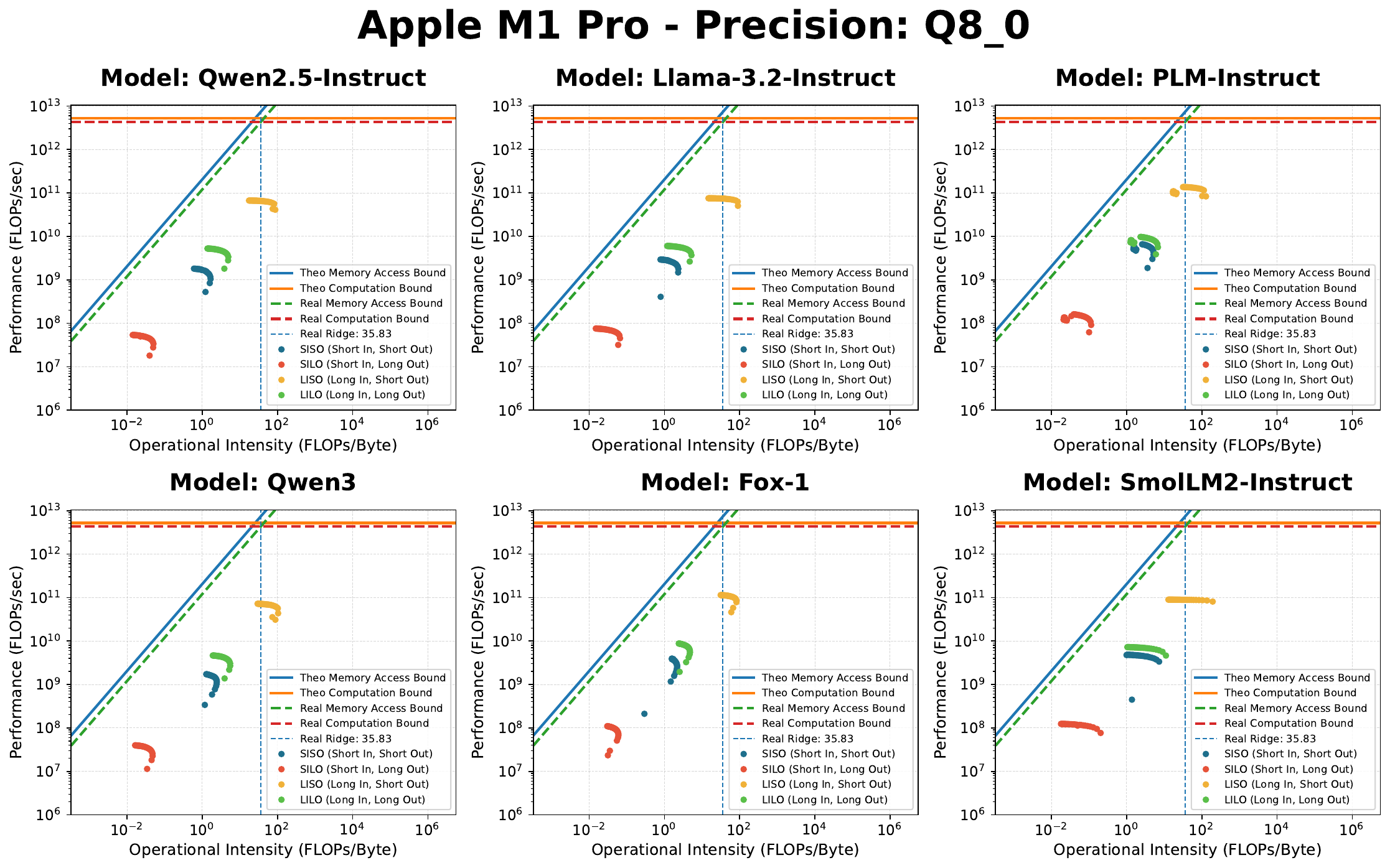}
    \caption{Performance profiles of diverse inference scenarios on Apple M1 Pro (Q8\_0). This set of roofline models highlights the performance of 8-bit quantized LLMs under four characteristic workloads: SISO, SILO, LISO, and LILO. The plots demonstrate that as output length increases (SILO/LILO), the operational intensity drops significantly, trapping the performance within the memory-bandwidth-limited regime. In contrast, LISO tasks with long input sequences move toward the compute-bound ceiling. The real ridge point remains constant at 35.83.}
\end{figure*}

\begin{figure*}[htbp]
    \centering
    \includegraphics[width=0.75\textwidth]{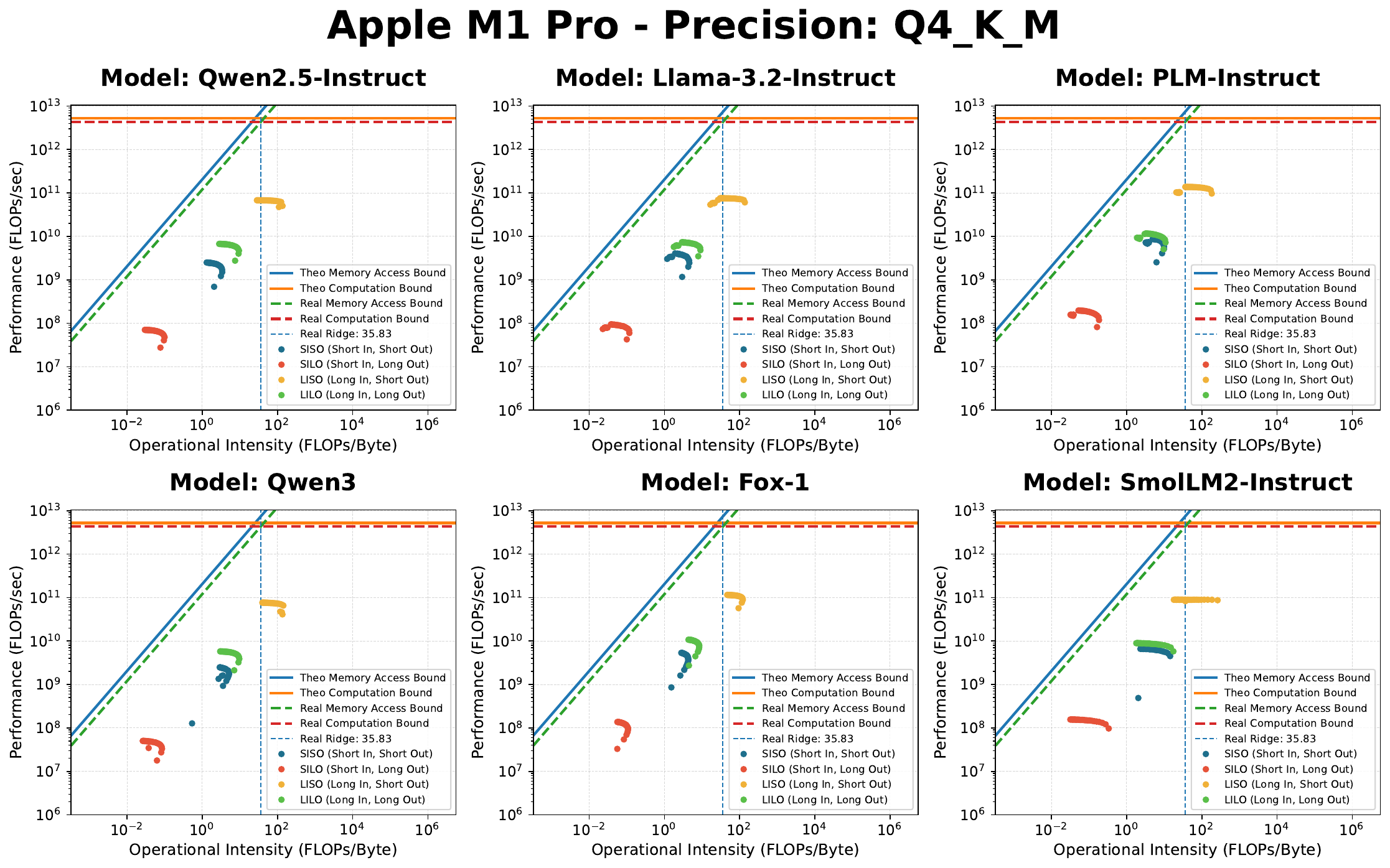}
    \caption{Performance profiles of diverse inference scenarios on Apple M1 Pro (Q4\_K\_M). This figure depicts the performance of 4-bit quantized models  across varied input and output context lengths. By categorizing tasks into SISO, SILO, LISO, and LILO clusters , the profiles reveal the impact of quantization on arithmetic intensity and throughput. Quantization allows models to achieve higher throughput relative to the real computation bound in prefill-dominant scenarios (e.g., LISO). The dashed vertical line marks the system's empirical ridge point at 35.83.}
\end{figure*}

\begin{figure*}[htbp]
    \centering
    \includegraphics[width=0.75\textwidth]{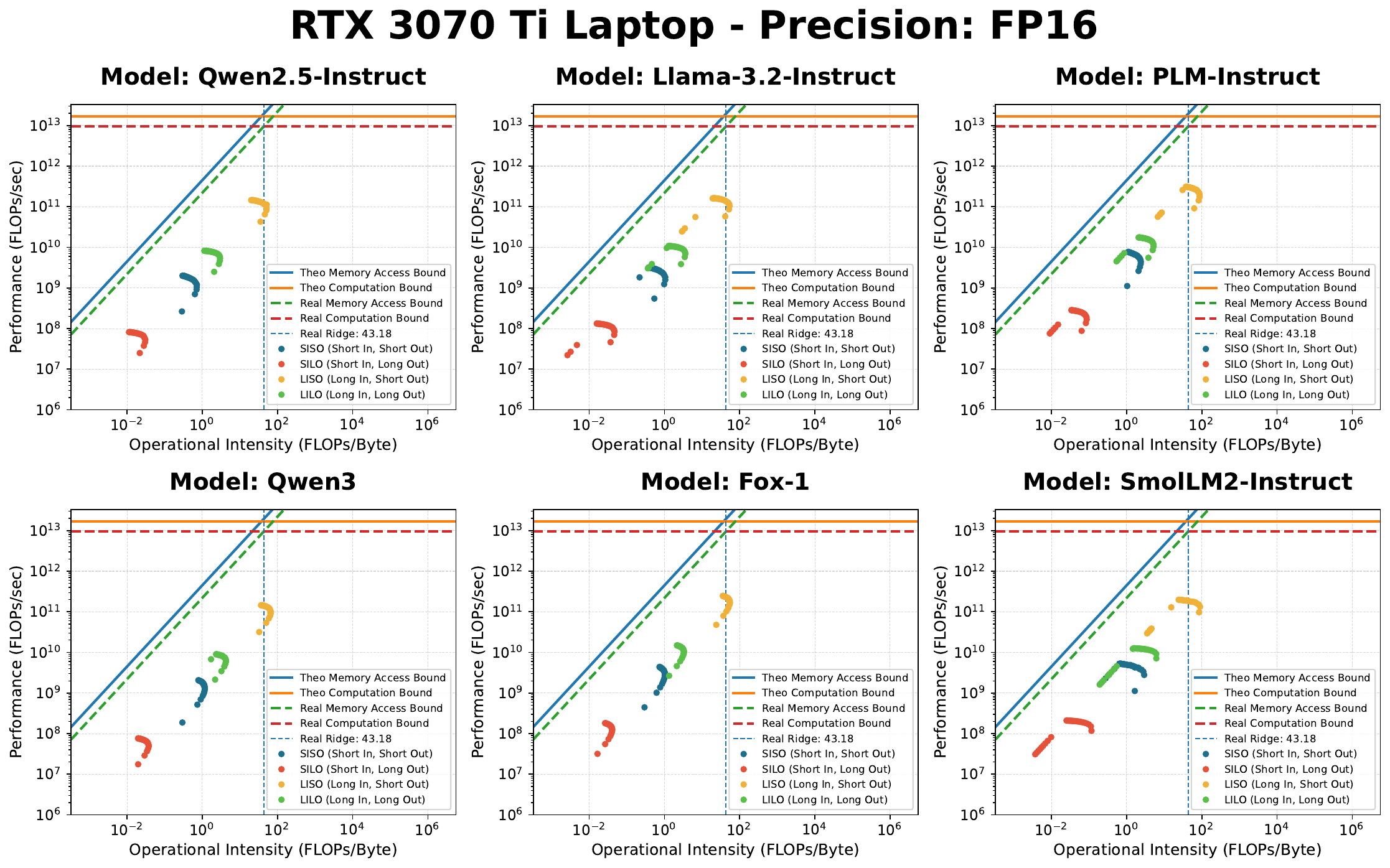}
    \caption{Performance profiles of diverse inference scenarios on RTX 3070 Ti Laptop (FP16). This figure evaluates the throughput ($FLOPs/sec$) across various models (e.g., Qwen2.5-Instruct, Llama-3.2-Instruct, and PLM-Instruct) under four distinct operational scenarios: SISO (Short In, Short Out), SILO (Short In, Long Out), LISO (Long In, Short Out), and LILO (Long In, Long Out). As operational intensity increases, tasks transition from the memory-bound region (e.g., SILO and LILO) toward the empirical computation bound, with prefill-heavy LISO scenarios exhibiting the highest arithmetic intensity. The hardware configuration exhibits a real ridge point of 43.18.}
\end{figure*}

\begin{figure*}[htbp]
    \centering
    \includegraphics[width=0.75\textwidth]{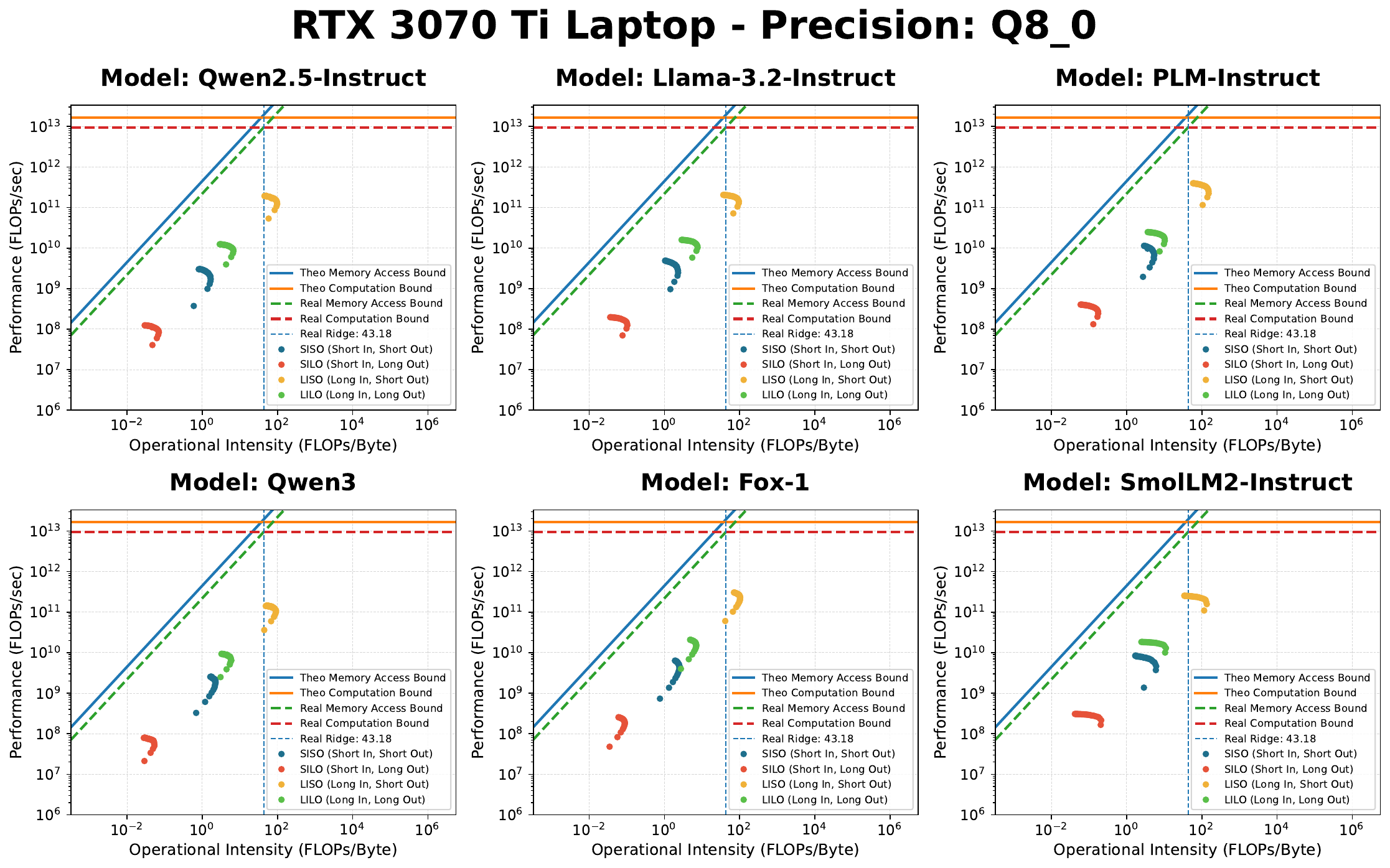}
    \caption{Performance profiles of diverse inference scenarios on RTX 3070 Ti Laptop (Q8\_0). These roofline models characterize the performance of 8-bit quantized LLMs (Q8\_0) under four characteristic workloads: SISO, SILO, LISO, and LILO. The results demonstrate that as output length increases (SILO/LILO), the operational intensity drops significantly, trapping performance within the memory-bandwidth-limited regime. In contrast, LISO tasks with long input sequences move toward the hardware's empirical computation ceiling. The real ridge point is consistently identified at 43.18.}
\end{figure*}

\begin{figure*}[htbp]
    \centering
    \includegraphics[width=0.75\textwidth]{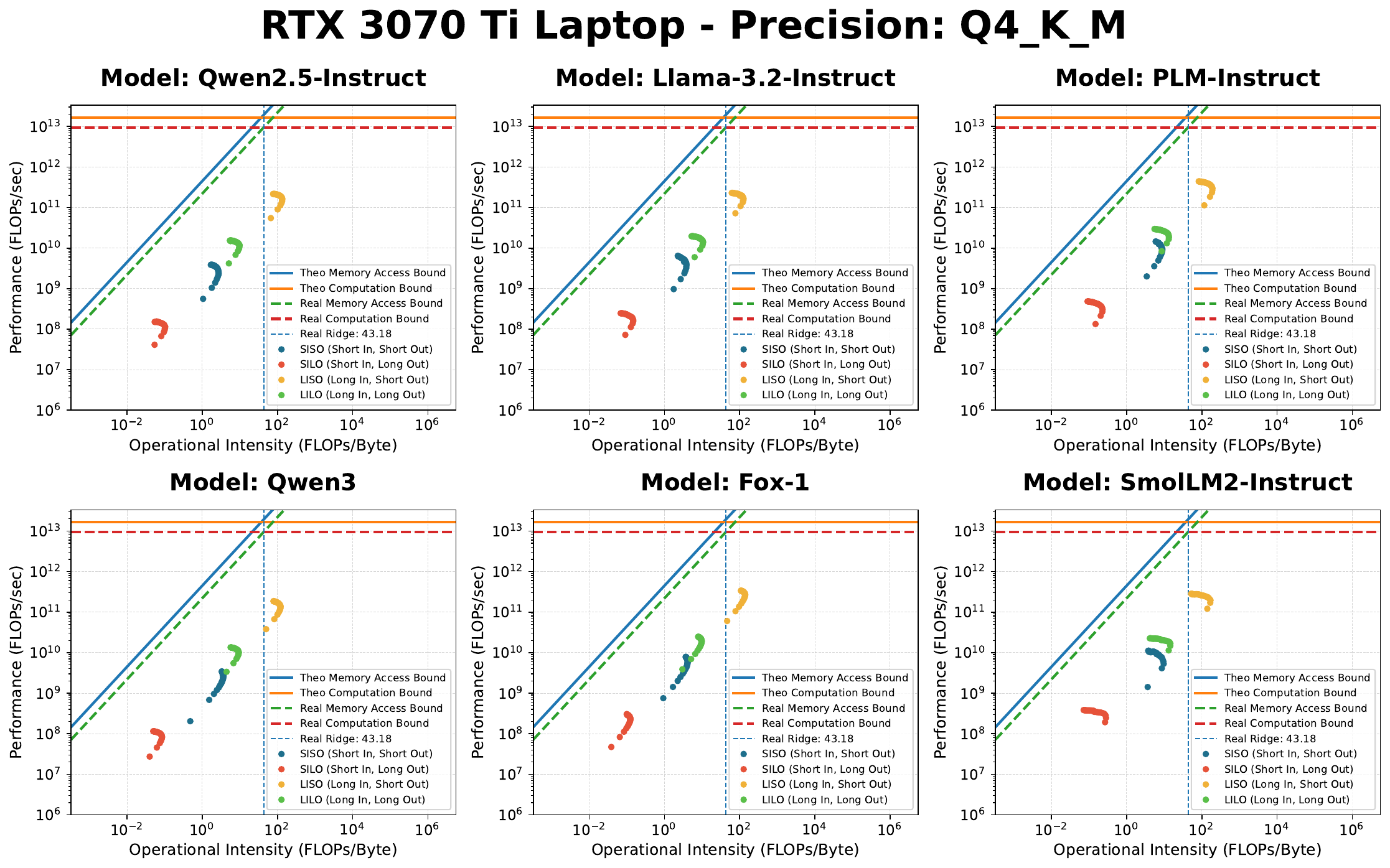}
    \caption{Performance profiles of diverse inference scenarios on RTX 3070 Ti Laptop (Q4\_K\_M). This set of profiles evaluates 4-bit quantized LLMs categorized into SISO, SILO, LISO, and LILO clusters to illustrate the impact of quantization on throughput across varied input/output lengths. Quantization shifts the compute-to-memory balance, allowing models to achieve higher throughput relative to the real computation bound in prefill-dominant scenarios (LISO). The dashed vertical line marks the system's empirical ridge point at 43.18.}
\end{figure*}

\begin{figure*}[htbp]
    \centering
    \includegraphics[width=0.75\textwidth]{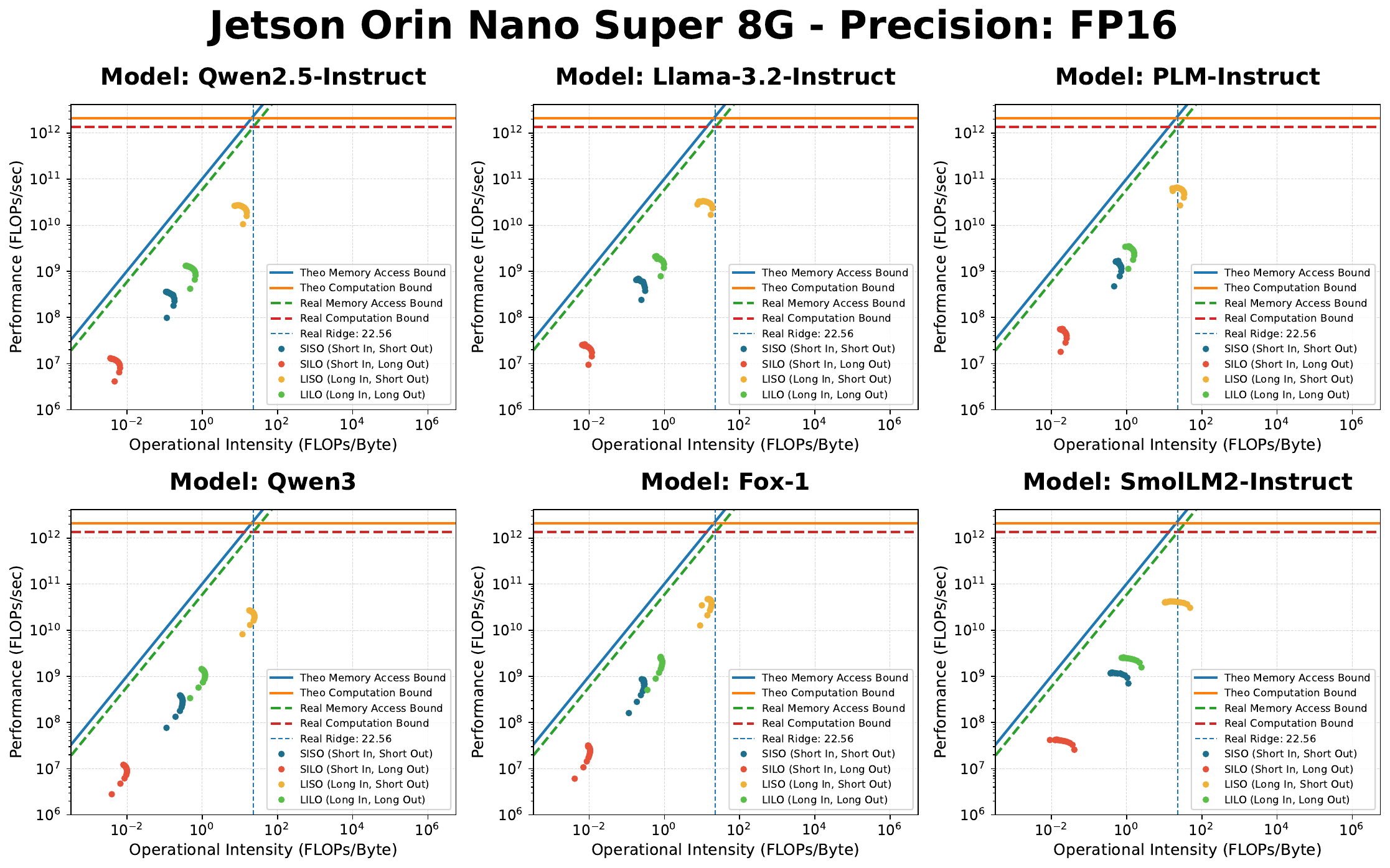}
    \caption{Performance profiles of diverse inference scenarios on Jetson Orin Nano Super 8G (FP16). This figure evaluates various models—including Qwen2.5-Instruct, PLM-Instruct, Qwen3, Fox-1, and SmolLM2-Instruct—across four distinct operational scenarios: SISO (Short In, Short Out) , SILO (Short In, Long Out) , LISO (Long In, Short Out) , and LILO (Long In, Long Out). The subplots illustrate how operational intensity determines the achieved throughput ($FLOPs/sec$). While decoding-heavy scenarios (SILO/LILO) remain memory-bound , prefill-dominant LISO tasks approach the real computation bound. The platform is characterized by a real ridge point of 22.56.}
\end{figure*}

\begin{figure*}[htbp]
    \centering
    \includegraphics[width=0.75\textwidth]{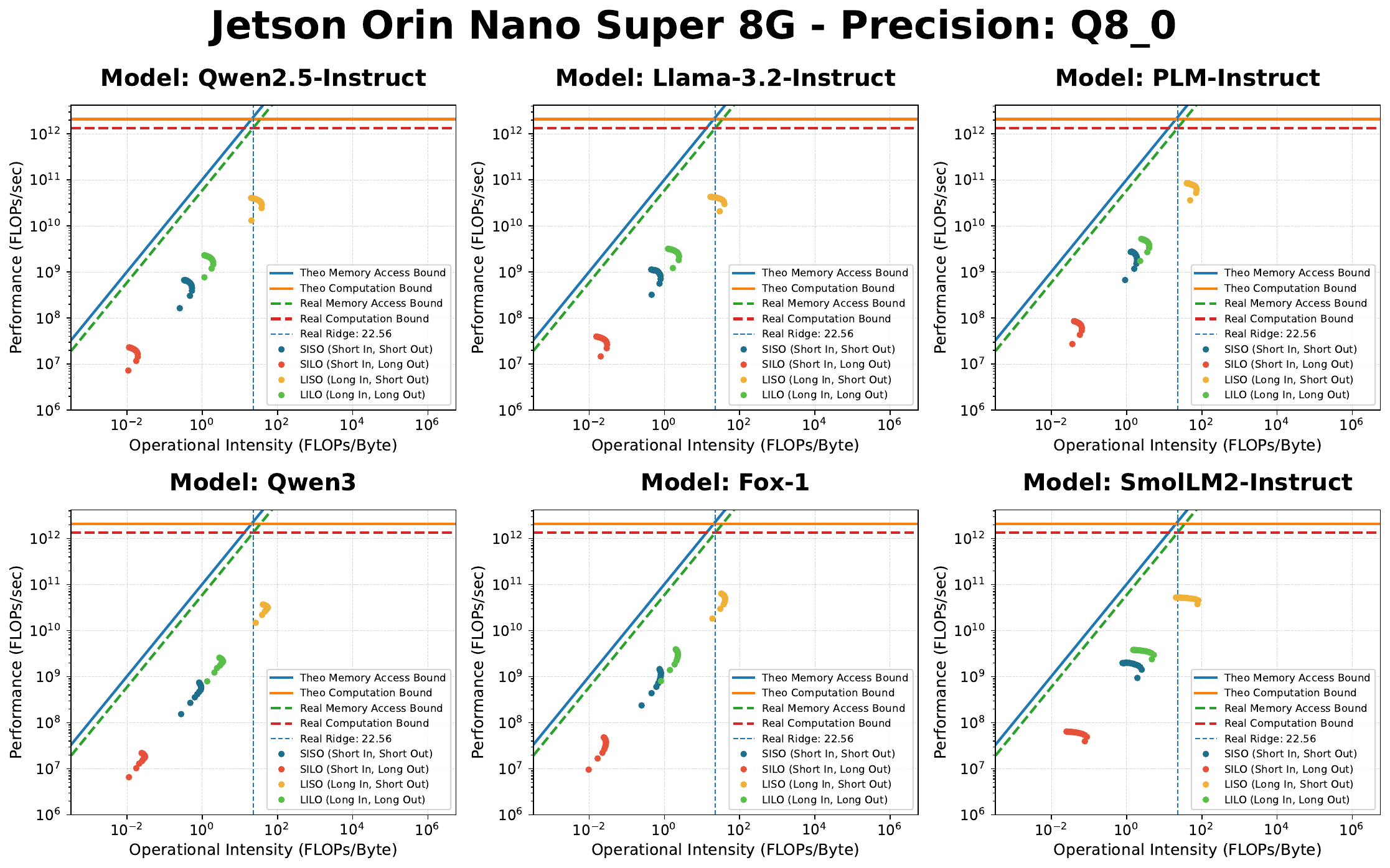}
    \caption{Performance profiles of diverse inference scenarios on Jetson Orin Nano Super 8G (Q8\_0). These roofline models highlight the performance of 8-bit quantized LLMs—including Qwen2.5-Instruct, Llama-3.2-Instruct, Qwen3, Fox-1, PLM-Instruct, and SmolLM2-Instruct—under four characteristic workloads: SISO, SILO, LISO, and LILO. The plots demonstrate that as output length increases (SILO/LILO), the operational intensity drops significantly, trapping the performance within the memory-bandwidth-limited regime. In contrast, LISO tasks with long input sequences move toward the hardware's empirical computation ceiling. Consistent with the platform profile, the real ridge point is consistently identified at 22.56.}
\end{figure*}

\begin{figure*}[htbp]
    \centering
    \includegraphics[width=0.75\textwidth]{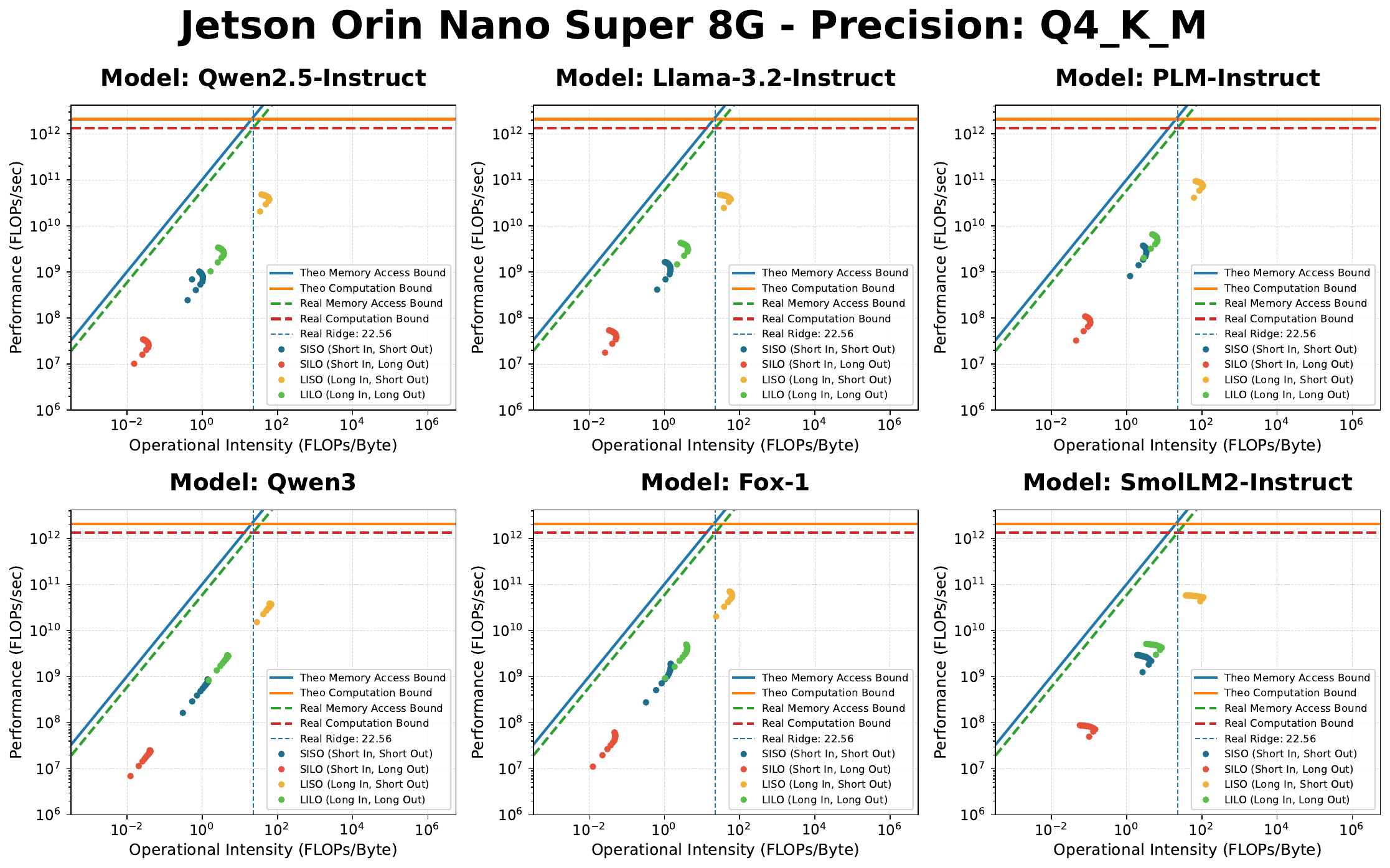}
    \caption{Performance profiles of diverse inference scenarios on Jetson Orin Nano Super 8G (Q4\_K\_M). This figure depicts the performance distribution of 4-bit quantized models—including Qwen2.5-Instruct, Llama-3.2-Instruct, PLM-Instruct, Qwen3, Fox-1, and SmolLM2-Instruct—across varied input and output context lengths. By categorizing tasks into SISO, SILO, LISO, and LILO clusters , the profiles reveal the impact of quantization on arithmetic intensity and throughput. Quantization allows models to achieve higher efficiency relative to the real computation bound, especially in LISO scenarios. The vertical dashed line indicates the system's empirical ridge point at 22.56.}
\end{figure*}

\begin{figure*}[htbp]
    \centering
    \includegraphics[width=0.75\textwidth]{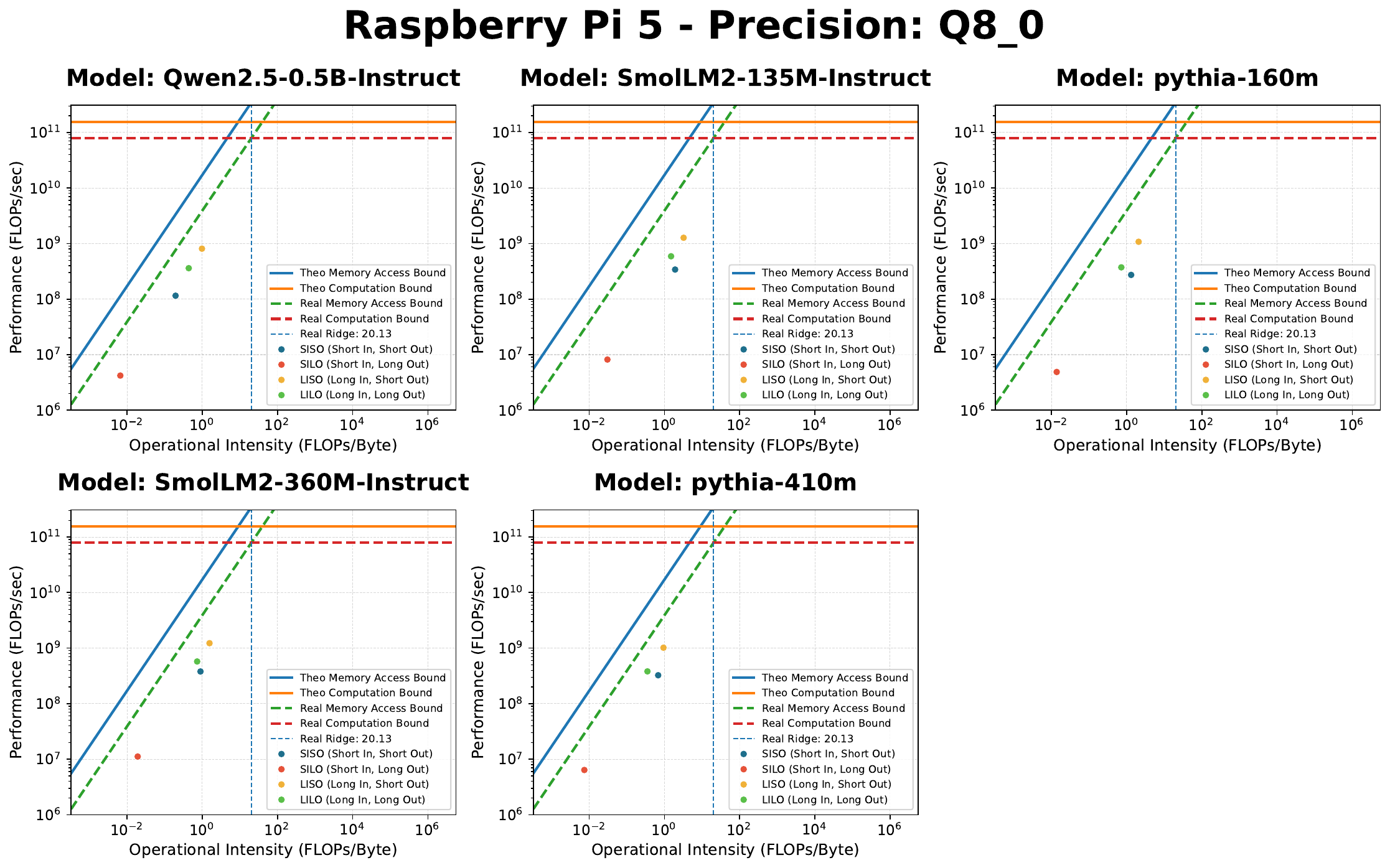}
    \caption{Performance profiles of diverse inference scenarios on Raspberry Pi 5 (Q8\_0). This figure characterizes the performance of 8-bit quantized lightweight LLMs—including SmolLM2-360M-Instruct and Pythia-410m—across four distinct operational workloads: SISO (Short In, Short Out), SILO (Short In, Long Out), LISO (Long In, Short Out), and LILO (Long In, Long Out). The subplots illustrate that in this resource-constrained environment, most scenarios are trapped within the memory-bandwidth-limited regime, particularly when output length increases (SILO/LILO). In contrast, LISO tasks with long input sequences exhibit significantly higher operational intensity, moving closer to the hardware's empirical computation ceiling. Theoretical and real bounds are plotted with a consistent real ridge point identified at 20.13.}
\end{figure*}

\end{document}